\documentclass[10pt,journal]{IEEEtran}
\usepackage{cite}

\usepackage{jabbrv}

\ifCLASSINFOpdf
  \usepackage[pdftex]{graphicx}
  \graphicspath{{./fig/}}
  \DeclareGraphicsExtensions{.pdf,.jpeg,.jpg,.png}
\else
\fi
\usepackage{amsmath}
\usepackage{amsthm}
\usepackage{amsfonts}
\usepackage{amssymb}
\usepackage{bbm}

\usepackage{url}

\usepackage{color}
\usepackage{soul}
\sethlcolor{white}
\soulregister\cite7
\soulregister\ref7
\soulregister\pageref7

\usepackage{multirow}
\usepackage{multicol}

\usepackage{color}
\usepackage{soul}

\theoremstyle{definition}

\theoremstyle{remark}

\theoremstyle{plain}

\usepackage[linesnumbered,ruled]{algorithm2e}

\usepackage{booktabs}

\usepackage[utf8]{inputenc}
\usepackage[T1]{fontenc}
\usepackage{makecell}

\begin{document}
%
\title{Diverse Instances-Weighting Ensemble based on Region Drift Disagreement for Concept Drift Adaptation}
%
%
%

\author{Anjin~Liu,~\IEEEmembership{Member,~IEEE,}~Jie~Lu,~\IEEEmembership{Fellow,~IEEE,}~and~Guangquan~Zhang
}

%
%

\markboth{IEEE TRANSACTIONS ON NEURAL NETWORKS AND LEARNING SYSTEMS, VOL. XX, NO. X, FEBRUARY 2019}
{Shell \MakeLowercase{\textit{et al.}}: Bare Demo of IEEEtran.cls for IEEE Journals}
%



\maketitle

\begin{abstract}
Concept drift refers to changes in the distribution of underlying data and is an inherent property of evolving data streams. Ensemble learning, with dynamic classifiers, has proved to be an efficient method of handling concept drift. However, the best way to create and maintain ensemble diversity with evolving streams is still a challenging problem. In contrast to estimating diversity via inputs, outputs, or classifier parameters, we propose a diversity measurement based on whether the ensemble members agree on the probability of a regional distribution change. In our method, estimations over regional distribution changes are used as instance weights. Constructing different region sets through different schemes will lead to different drift estimation results, thereby creating diversity. The classifiers that disagree the most are selected to maximize diversity. Accordingly, an instance-based ensemble learning algorithm, called the diverse instance weighting ensemble (DiwE), is developed to address concept drift for data stream classification problems. Evaluations of various synthetic and real-world data stream benchmarks show the effectiveness and advantages of the proposed algorithm.
\end{abstract}

\begin{IEEEkeywords}
concept drift, data stream, ensemble learning, classification
\end{IEEEkeywords}

%
\IEEEpeerreviewmaketitle

%
%
%
%






\section{Introduction}
\label{sce-in}

\IEEEPARstart{C}{onventional} machine learning methods assume learning environments are stationary, that is, the testing data has the same data generation distribution as the training data. However, this assumption is substantially undermined in the context of the Internet of Things and Big Data \cite{Viktor:SAMkNN, Gama:survey, Liu:Survey}, where data distributions can easily change over time. A data distribution is just a reflection of how frequently a real-world concept appears in some data. Hence, this phenomenon of changing distributions has been termed concept drift and, with today's advancing data streams, learning in its presence has become an important topic of research. A typical symptom of concept drift is a shift in the decision boundary of a classification, which reduces prediction accuracy \cite{Liu:Survey}. For example, consider a user preference prediction or fraud detection task on streaming data. The performance of a static predictor trained on historical data will inevitably degrade over time because the nature of personal preferences or fraudulent attacks is that they are always evolving \cite{Harel:Gradual}. As concepts change, new data may no longer conform to old patterns \cite{Lu:AI1}, which negatively impacts subsequent data analysis tasks \cite{Lu:AI2}. More importantly, these changes may be barely perceptible in real-world scenarios. For this reason, a continuous learning system will vigilantly monitor concept drift and adapt to it quickly, rather than assuming a learning environment is stationary.

Ensemble algorithms are useful for data stream learning as they can be integrated with drift detection algorithms and updated dynamically \cite{Bifet:Survey}. Two comprehensive surveys on data stream ensemble learning \cite{Bifet:Survey, Wozniak:EnsembleSurvey} have cataloged the advantages and disadvantages of current ensemble learning algorithms. Both point out that dedicated diversity measurements for data stream classifier ensembles are a worthwhile direction of research. Additionally, the literature shows that most existing ensemble diversity measurements are based on the outputs of classifiers \cite{Brzezinski:Diversity} - for example, the Kohavi-Wolpert variance, double fault, the interrater agreement, Yule’s Q statistic coincident failure diversity, etc. \cite{Brzezinski:Diversity,Kuncheva:2004Book}. Such diversity is created via input manipulation, output manipulation, base learner manipulation, or heterogeneous base learners \cite{Bifet:Survey}. As reported in \cite{Brzezinski:Diversity}, existing ensemble diversity measurements are designed for static learning, but there have been no proposals for an ensemble diversity measurement specifically for evolving data streams.

To the best of our knowledge, no established ensemble diversity measurement can reflect disagreements between the base learners as to whether a concept drift has occurred or not. We propose that an ensemble diversity measurement for changing data streams should be able to address this issue since it can help the ensemble system to  select the most appropriate ensemble members (ensemblers) for the final prediction. High diversity between two base learners indicates that one learner has found a drift while the other has not. Since every base learner updates itself with its own rule, different opinions on the existence of drift will result in different update statuses. This creates diversity in itself and is the inspiration behind our thinking. To implement this idea, we propose to detect drift using different change detection settings, and adjust instances' weights to adapt to drift. In other words, drift detection works at the instance level, where a data instance is considered less important if it is located in a region of drift \cite{Liu:LDD_DSDA, Liu:PR}. A region drift disagreement index is proposed as a tool to measure ensemble diversity and to select the most appropriate ensemblers to vote for in the final prediction.

The predictive accuracy of current concept drift detection-adaptation strategies is highly reliant on accurate change detection and a low false positive rate. Missing a change or raising a false alarm may impair their overall performance. In contrast, in an innovative way, our stance is that whether concept drift is present or not is uncertain. Under this assumption, it is reasonable that different change detection tests could have different change detection results. Therefore, instead of updating the learning models whenever an ensemble detects a drift, we have opted for a voting strategy where the ensemblers with the most controversial change detection results act as the voting representatives. In broad terms, the innovation here is to leverage the base learner's update mechanism as a way to create ensemble diversity. Assigning different change detection settings to each ensembler will give rise to disagreement and then the members with the most controversial results can be selected to maximize diversity.

The main contribution of this paper is a diverse instance-weighting ensemble algorithm (DiwE) based on region drift disagreement for concept drift adaptation. DiwE consists of an instance weighting method to incrementally adjust sample weights and an ensemble diversity measurement to select the ensemblers. By defining different schemes for constructing the region sets, DiwE can dynamically change the weights of instances according to newly-emerging concepts and, further, can select the combination of ensemblers with the highest diversity. Compared to other concept drift adaptation ensemble algorithms, DiwE has the following advantages:

\begin{enumerate}
    \item It can incrementally adapt instance weights based on the estimated risk of region drift and pass this information to learning models before a drift becomes statistically significant.
    \item It can dynamically select different ensemblers in different concept drift situations and automatically adapt to the diversity present, which is not possible with existing methods.
\end{enumerate}

The rest of this paper is organized as follows. The definition of concept drift is provided in Section \ref{s:related_works} along with a review of relevant works. Section \ref{s:DiwE} formally describes the proposed DiwE algorithm for concept drift adaptation and its associated adjuncts, including the region drift disagreement index and procedure for selecting the ensembler with the maximum disagreement in region drift. Section \ref{s:DiwE_evaluation} presents the evaluation of DiwE using benchmarks that include artificial streams with known drift characteristics and highly-referenced real-world applications. Section \ref{s:VI} concludes this study with a discussion on future work.


\section{Preliminary and Related Works}
\label{s:related_works}
This section begins by introducing the preliminaries, definitions, and types of concept drift. The state-of-the-art ensemble-based algorithms for handling concept drift are then categorized based on their drift detection and adaptation strategies.

\subsection{The definition of concept drift}
\label{ss:concept_drift_definitions}
Concept drift is caused by variations in the distribution of data \cite{Gama:survey},  which, in turn, leads to a disparity between the training samples and the data streams associated with non-stationary learning environments \cite{Jose:ConceptShift}.
Denote the feature space as $\mathcal{X}\subseteq \mathbb{R}^n$, where $n$ is the dimensionality of the feature space. A data instance $d_t=(X_t,y_t)$ is a pair of feature vectors $X_t$ and a label $y_t$, $y_t\in\{y_1,\ldots, y_c\}$, where $c$ is the number of classes. A data stream can then be represented as an infinite sequence of data instances denoted as $D$. A concept drift occurs at time $t+1$ if the joint probability of $X$ and $y$ changes, that is, $\exists X \in \mathcal{X},~t\in \{1,\ldots,m\}:p_{0, t}(X,y)\neq p_{t+1, m}(X,y)$ \cite{Lu:AI2, Liu:LDD_DSDA, Gama:survey, song2019fuzzy}, where $m$ is the number of total available data instance. In this paper, time is considered to be discrete.

If we further decompose $p(X,y)$, we have $p(X,y)=p(y|X)\cdot p(X)$. In considering problems that use $X$ to infer $y$, concept drift is generally divided into two sub-research topics \cite{Jose:ConceptShift}:
\begin{itemize}
    \item Covariate shift focuses on the drift in $p(X)$ while $p(y|X)$ remains unchanged. This is known as $virtual$ $drift$ \cite{Polikar:Survey, Gama:survey, Wozniak:Survey}.
    \item Concept shift focuses on the drift in $p(y|X)$ while $p(X)$ remains unchanged. This is most commonly referred to as concept drift, sometimes called $actual$ $drift$ or $decision$ $boundary$ $drift$ \cite{Polikar:Survey, Gama:survey, Wozniak:Survey}.
\end{itemize}

By this definition, a drift consists of both the time information $t$ and the location information $X$ in the feature space. Note that $p(X)$ and $p(y|X)$ are not the only elements affected by $p(X,y)$ drift. The prior probabilities of classes $p(y)$ and the class conditional probabilities $p(X|y)$ may also change, which could lead to $p(y|X)$ changing, again affecting \cite{Gama:survey}. This issue is the next challenge to consider in concept drift learning, i.e., understanding the reason for the drift \cite{Liu:Survey}.

Tsymbal et al. \cite{T:local_IF} first defined region drift as changes in the concepts and the data distributions at the instance level rather than at the dataset level. They noted that lazy learning is a good option for addressing region drift problems. Other related studies that address region drift with decision tree models \cite{Gama:Local_FIMT-DD_JNL,Gama:Local,Bifet:HAT-ADWIN_HWT,Bifet:HAT_local_variant} have also demonstrated good results. However, decision tree models require a minimum number of instances to perform their splitting algorithm. For example, the CVFDT algorithm \cite{Hulten:CVFDT} normally requires observation of 200 data instances before attempting to split the nodes. If a region drift occurs within those 200 data instances, a tree node will be updated before splitting, which means no region drift will be identified in that node \cite{Liu:PR}. In \cite{Liu:LDD_DSDA,Liu:PR}, the authors proposed quantifying the discrepancies in regional density with a metric they called the local drift degree \cite{Liu:LDD_DSDA}. These discrepancies are accumulated to determine whether the overall change is sufficiently significant to report concept drift \cite{Liu:PR}. These studies, like many other drift detection algorithms \cite{I:HDDM,Alippi:LSDD-CDT,Alippi:HCDTs}, require prior knowledge to organize the data samples in a stream into time-window sample sets. If a drift is confirmed, the old-time window is replaced with the latest time window, but the non-drift information in the old window is not reused. An empirical method for selecting the region size has been developed, based on a metric called the information granularity indicator. But how a region of concept drift should be defined at the instance level, along with the size of that region, are both still unsolved.

\subsection{Ensemble-based algorithms for concept drift handling}
\label{ss:ensemble-concept-drift}

The research on handling concept drift mainly covers drift detection and drift adaptation. The aim of drift detection is to determine the time at which a drift occurs and notify a learning model to update itself. Drift adaptation focuses on how to update a learning model with the least effort to achieve the best learning results.

An ensemble is a set of individual classifiers whose predictions are combined to predict (e.g., classify) new incoming instances. This is considered to be one of the most promising research directions for intelligent data stream analysis \cite{Wozniak:EnsembleSurvey}. Ensembles for concept drift detection seek to improve the precision of identifying a change (drift) and reduce false alarm rates. These types of algorithms for drift detection are also known as multiple hypothesis testing \cite{Liu:Survey}. Ensembles for concept drift adaptation aim to improve overall prediction accuracy by decomposing a complex learning problem into easier sub-problems \cite{Wozniak:EnsembleSurvey,Bifet:Survey}. Ensemble algorithms are efficient at drift adaptation because they can easily incorporate dynamic updates, such as selective removal or addition of classifiers \cite{Bifet:Survey}. The types of algorithms have two modes – online-mode and chunk-mode – and fall into two categories: the active category and the passive category \cite{Polikar:Survey}. Online mode algorithms process data instances one-by-one. Chunk mode algorithms process data instances in fixed batch sizes (i.e., chunks) \cite{Wozniak:EnsembleSurvey}. 

The active category relies on change detection tests to trigger the adaptation process. The tests inspect the features extracted from the data generation process and/or from an analysis of the classification errors (evaluated over labeled samples) \cite{Polikar:Survey}. Any detected changes are then accommodated by either updating or retraining the classifier(s). Of course, to perform well, changes need to be detected promptly and false positive rates need to be controlled \cite{Alippi:HCDTs, Boracchi:QTree, Boracchi:UniHist}. 
Unlike the active category, the passive category does not actively detect drift as new data arrives. Rather it simply accepts that the underlying data distributions may change at any time and at any rate \cite{Polikar:Survey}. To accommodate this uncertainty, the model is adapted every time new data arrives.


From a drift detection perspective, ensemble algorithms fall into two categories \cite{Lu:AI1,Polikar:Survey,Viktor:SAMkNN}: active drift detection  with an adaptive ensemble; or a passive ensemble with a forgetting mechanism.
Algorithms like ADWIN-ARF \cite{Bifet:ADWIN_ARF} and leverage bagging \cite{Bifet:ADWIN_LeverageBagging} fall into the first category. With these, the ensemble actively searches for concept drift and builds new ensemblers if a drift is detected. 
In the second category, the ensemblers are built without considering the conflicts between concepts, and the base learners are built according to a predefined time frame without explicit drift detection. Example algorithms in this second category include DWM \cite{K:DWM}, Learner++.NSE \cite{Polikar:NSE}, AUE1 AUE2 \cite{Brzezinski:AUE2}, and OnlineAUE \cite{Brzezinski:OAUE}. These algorithms attempt to learn drift incrementally with each new piece of arriving data, eliminating old ensemblers through a forgetting mechanism \cite{Xin:CBCE}. The major difference between the two categories is whether the ensemble algorithm contains an explicit drift detection method. Other interesting research about how ensemble diversity may affect drift adaptation is discussed in \cite{Minku:DiversityImpact, Minku:DDD}.

Class imbalance is another problem with concept drift. The class imbalance problem in sequential learning has garnered increasing attention from researchers in various application domains. The two most prominent solutions are ensemble of subset online sequential extreme learning machine (ESOS-ELM) \cite{mirza2015ensemble} and meta-cognitive online sequential extreme learning machine (MOS-ELM) \cite{mirza2016meta}. A study of online class imbalance learning with concept drift can be found in \cite{Xin:ImbalanceSurvey}.


\subsection{Ensemble diversity measurement}
\label{ss:Ensemble-diversity-measurement}
Ensemble diversity seeks to quantitatively analyze the dissimilarity between a set of individual classifiers. As illustrated in \cite{Bifet:Survey}, the importance of ensemble diversity can be intuitively explained using the anthropomorphic example of a group of individuals with different knowledge backgrounds who need to make decisions together. If the group had the same knowledge backgrounds, they would not be able to think about the problem from different angles, while a diverse group of individuals is more capable of lateral thinking.

Moreover, some correlations between accuracy and specific diversity measurements have been found in special cases \cite{Kuncheva:DiversityMeasureMachineLearning, Minku:DiversityImpact, Minku:DDD}. For example, Minku et al. \cite{Minku:DiversityImpact} argue that, before a drift, ensembles with less diversity have lower test errors, while, after a drift, maintaining highly diverse ensembles could result in lower test errors. The authors also find that diversity is beneficial for reducing the errors caused by a drift, but it does not speed up recovery from a drift over the long term \cite{Minku:DiversityImpact}. Their argument is well supported by comprehensive evaluations. However, theoretical guarantees for more general cases are yet to be discovered \cite{Bifet:Survey}. In addition, since there is no generally-accepted definition of diversity, a way of proving the correlations between accuracy and diversity is still not clear \cite{Kuncheva:2004Book}.

Most ensemble algorithms are accompanied by strategies to create diversity, even if the strategies are not part of the core algorithm \cite{Bifet:Survey}. Only a few studies \cite{Bifet:Survey, Brzezinski:Diversity, Minku:DiversityImpact, Minku:DDD} have been undertaken to devise specific diversity measurements and their properties for ensemble learning with data streams. 
Brzezinski and Stefanowski’s approach \cite{Brzezinski:Diversity} is to visualize a diversity measurement over time and use those values as complementary information to the data stream. 
Overall, diversity creation methods can be divided into four broad categories: input manipulation, output manipulation, base learner manipulation, and heterogeneous base learners \cite{Bifet:Survey}. However, there is still a gap in how to define ensemble diversity for concept drift detection and adaptation \cite{Wozniak:EnsembleSurvey}.


\section{A Diverse Instance Weighting Ensemble Algorithm via Drift Risk Estimation on Different Region Sets}
\label{s:DiwE}

Our proposal for diverse instance weighting is based on region drift estimation, where a region is defined as an $n$-ball, i.e., a ball in an $n$-dimensional Euclidean feature space. The intuition behind the idea is to estimate concept drift at the instance level rather than across the entire feature space. Because detecting drift across the entire feature space has the risk that sub-spaces with insignificant changes may be combined, resulting in substantially fewer discrepancies in density\cite{A:Dilute}. In general, the proposed method choose different region sizes in a range to perform ensemble learning. 
And the ensemblers with the highest regional drift disagreement are selected to perform the final voting.
In this paper, we assume that data instances arrive independently identically distributed (i.i.d.) in an online mode.

\subsection{A phi-level region set}
\label{ss:region-set}

\subsubsection{The definition of a phi-level region set}

In region drift detection, the first critical problem is how to define a region and how to choose the region size. 
The selection of region size affects the sensitivity of drift detection \cite{Liu:LDD_DSDA,Liu:PR}. Large regions are not sensitive to local drifts, but they are robust to noise. Conversely, small regions are sensitive to local drifts, but may be affected by noise.
Setting a specific distance for constructing a region is one option; however, using a fixed distance has several drawbacks when dealing with a high-dimensional feature space, arbitrary shapes, or distributions with high-density variations between different regions. For example, sparse regions may have no data instances, while dense regions may have too many. Similar problems can occur when using kernel density estimation. Therefore, we have considered region size from a probability perspective rather than a geometric view. In other words, the region size should be determined by the relative proportion of the region sample.

A $\phi$-level region set $\mathbb{B}_{\phi}=\{\mathcal{B}_1^{\phi},\ldots,\mathcal{B}_m^{\phi}\}$ is defined as a set of $n$-balls of the feature space $\mathcal{X}$, $\mathcal{B}_i^{\phi}(d_i,\varepsilon_i^\phi)\subset \mathcal{X}$, where $d_i$ is the core data instance of the region, $\varepsilon_i^\phi$ is the radius, and $m$ is the number of available data instances. The sample proportion in each region is equal to $\phi$, namely $\mathrm{Pr}(X_j\in\mathcal{B}_i^\phi )=\phi$. The parameter $\phi$ ranges between 0 and 1, and determines the radius $\varepsilon_i^\phi$. 

\begin{figure}[!ht]
    \centering
    \includegraphics[scale=0.5]{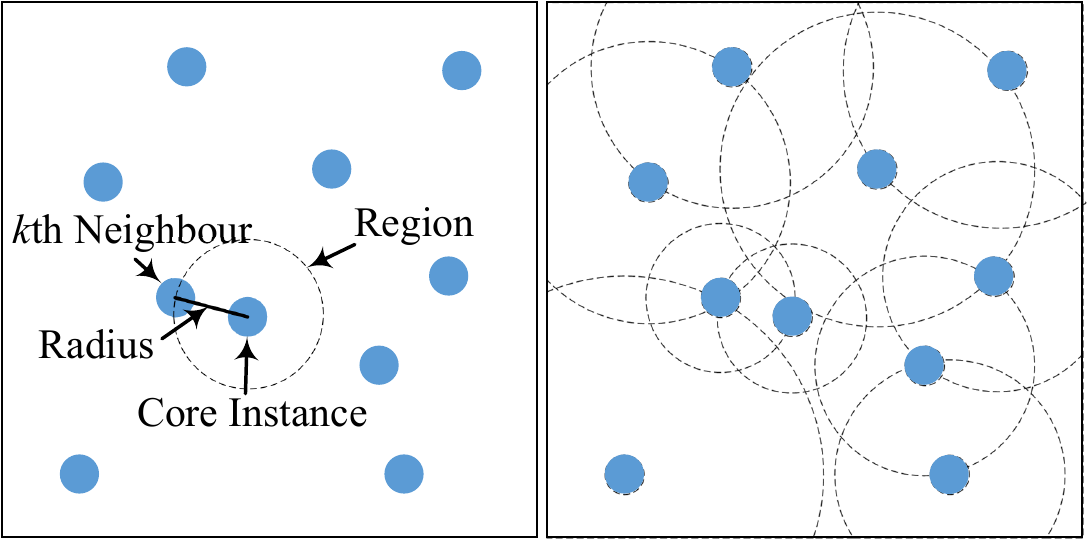}
    \caption{An illustration of a $\phi$-level region set. The left subfigure shows the components of a region. The right shows the $\phi$-level region set. In this example, the data size is $m=10$, and the empirical region sample proportion is $\hat{\phi}=0.2$. In a real-world scenario, m would need to be much larger to build a region set.}
    \label{fig:region}
\end{figure}

To implement a $\phi$-level region set for computation, we opted a $k$-nearest neighbor-based region construction method. Consider a data instance $d_t$ in a non-drift period $t\in \mathbb{Z}_{\leq m}^+$ as the center and a distance $\varepsilon_t^\phi$ as the radius. The empirical $\hat{\phi}$ of the region can be estimated by the number of instances in the region divided by the total number of instances, denoted as $\hat{\phi}=m^{\mathcal{B}_t^{\phi}}/m$ , where $m^{\mathcal{B}_t^{\phi}}$ is the cardinality of $D_{1,m}^{\mathcal{B}_t^{\phi}}$, $D_{1,m}^{\mathcal{B}_t^{\phi}}=\{d_i: i\in \mathbb{Z}_{\leq m}^+ ~ and ~ \|X_i- X_t \|<\varepsilon_t^\phi \}$, and $\|X_i- X_t \|$ denotes the Euclidean distance between the features of $d_i$ and $d_t$. The interval or radius of a region $\mathcal{B}_t^{\phi}$ is determined by the distance between the data instance $d_t$ and its $k^{th}$ nearest neighbor, denoted as
\begin{equation}\label{equ:knn_distance}
    \varepsilon_t^\phi=\underset{k\in \mathbb{Z}^+, k\leq\phi\cdot m}{\max}(\|X_t-X_k\|).
\end{equation}
As such, a larger $k$ value implies a larger region size, which is less sensitive to small discrepancies in density.
Figure \ref{fig:region} illustrates the components of a region and the region set in a 2-dimensional space.

\subsubsection{The minimum sample size to initialize a region set}
\label{subsub_m0}
With sufficient data, a $\phi$-level region can be constructed by choosing $k=\lceil \phi \cdot m\rceil$ and calculating the distance between $d_t$ and its $k^{th}$ nearest neighbor as the radius. According to Box et al. \cite{box:sampleSize}, the distribution of $m^{\mathcal{B}_t^{\phi}}$ and the region sample proportion $\phi$ are approximately normal for large values of $m$. This result follows the central limit theorem. The mean and variance for an approximately normal distribution of $m^{\mathcal{B}_t^{\phi}}$ is $m\cdot\phi$ and $m\cdot\phi\cdot(1-\phi)$, respectively, which is identical to the mean and variance of a binomial distribution. Similarly, the mean and variance for an approximately normal distribution of the sample proportion is $\phi$ and $\phi\cdot(1-\phi)/m$, respectively. However, because a normal approximation is not accurate for small values of $m$, a good rule of thumb is to use the normal approximation only if $m\cdot\phi>10$ and $m\cdot(1-\phi)>10$ \cite{box:sampleSize}. In other words, the number of available data $m$ should be greater than $\max\{ 10/\phi, 10/(1-\phi)\}$. For example, if we want to create a region set with $\phi=0.2$, the minimum number of data instances we need is $\max\{ 10/0.2, 10/(1-0.2) \}=50.$

\subsection{An incremental instance-weighting function}
\label{ss:incr-weighting}

\subsubsection{The intuition behind diverse instance weighting}

The fundamental idea of instance weighting is that, if there is no concept drift in the streaming data, the next incoming $\tau$ data instances, whether or not they are located in a region, can be considered as a Bernoulli process. The set of data instances that enter the region $\mathcal{B}_t^{\phi}$ from the next continuously arriving $\tau$ samples is denoted as $D_{t+1,t+\tau}^{\mathcal{B}_t^{\phi}}=\{d_i: i\in\mathbb{Z}_{t+1\leq i\leq t+\tau}^+ ~ and ~ \|X_i- X_t \|<\varepsilon_t^\phi \}$. If no drift exists between the time points $\{1,\ldots,t+\tau\}$, we have a random variable, the cardinality of 
$m_{t+1,t+\tau}^{\mathcal{B}_t^{\phi}}=|D_{t+1,t+\tau}^{\mathcal{B}_t^{\phi}}|$, follows the binomial distribution, denoted as
\begin{equation}\label{equ:hits_follow_binomial}
    m_{t+1,t+\tau}^{\mathcal{B}_t^{\phi}}\sim B(\tau, \phi).
\end{equation}
%
Therefore, the probability of observing a number $\hat{m}_{t+1,t+\tau}^{\mathcal{B}_t^{\phi}}$ at time $t+\tau$ can be calculated as
\begin{equation}\label{equ:p_of_u}
    \mathrm{Pr}(m_{t+1,t+\tau}^{\mathcal{B}_t^{\phi}}=\hat{m}_{t+1,t+\tau}^{\mathcal{B}_t^{\phi}})=b(\hat{u}_t,\tau,\phi),
\end{equation}
where $b(\hat{u}_t,\tau,\phi)$ is the probability mass function of the binomial distribution. If $\mathrm{Pr}(m_{t+1,t+\tau}^{\mathcal{B}_t^{\phi}}\leq \hat{m}_{t+1,t+\tau}^{\mathcal{B}_t^{\phi}} )<\alpha$, the event has a small probability of occurring, and that drift level is reported for data instance $d_t$, where $\alpha=1-\mathrm{pValue}$ controls the sensitivity to drift.

To implement this approach incrementally, we have placed the focus on calculating $F(\hat{m}_{t+1,t+\tau}^{\mathcal{B}_t^{\phi}}=0,\tau,\phi)$, which is the probability that no other data instance will be in region $\mathcal{B}_t^{\phi}$ in the next $\{t+1,\ldots,t+\tau\}$ period. This online region drift condition can be rewritten as
\begin{equation}\label{equ:p_zero_cum}
    \begin{split}
        F(0,\tau,\phi)<\alpha \Rightarrow (1-\phi)^\tau< \alpha.
    \end{split}
\end{equation}
Then, we have $F_{\tau+1}(0,\tau+1,\phi)=F_{\tau}(0,\tau,\phi)\cdot(1-\phi)$, if $\|d_{\tau+1}-d_{\tau}\|>\varepsilon_{\tau}^\phi$, which is used as the weighting function.

The region $\mathcal{B}_t^{\phi}$ will be updated if it exists a time $\tau_0$ $(t<\tau_0<t +\tau)$ that satisfies the following conditions:
\begin{equation}\label{equ:update}
    \exists\tau_0\in \mathbb{Z}^{+} s.t. ||X_{t+\tau_0}-X_t||\leq \varepsilon_t^{\phi}~and~F(0,\tau_0,\phi)\geq\alpha.
\end{equation}
This ensures that the radius of the regions become more accurate as the amount of available data increases.

Essentially this means that each region of drift is detected based on the sequence of data instances arriving from time point 1 to time point $t+\tau$, rather than using the data instances in a region. For example, if a data instance arrives at time point $t=500$, denoted as $d_{t_{500}}$, region $\mathcal{B}^{\phi=0.1}_{t_{500}}$ will be built based on the buffered 499 data instances $\{d_{t_1},\ldots,d_{t_{499}} \}$. Setting the region set parameter to $\phi=0.1$ and the drift significance level at $\alpha=0.01$, $d_{t_{500}}$ will be reported as a drift instance only if no data instance are located in region $\mathcal{B}_{t_{500}}^{\phi=0.1}$ over the next $\tau=\lceil \log_{(1-\phi)} \alpha \rceil=\lceil \log_{(0.9)} 0.01 \rceil=44$ instances, which is the period $t\in\{501, \ldots, 545\}$.

In addition, the region $\mathcal{B}^{\phi=0.1}_{t_{500}}$ and the time counter $\tau$ will be reset if a new coming instance locates in the neighbourhood of $d_{t_{500}}$. This ensures the tested neighbourhoods and time periods are independent from previous tests so that DiwE will not perform repeated hypothesis tests. For example, if a data instance arrived at $t=520$ located in the neighbour of $d_{t_{500}}$, the $\mathcal{B}^{\phi=0.1}_{t_{500}}$ will be rebuilt based on the buffered $500 + 20=520$ data instances, and the $\tau$ will be reset to $0$. Then the $d_{t_{500}}$ will be reported as a drift instance only if no data instance are located in region $\mathcal{B}_{t_{500}}^{\phi=0.1}$ for the period $t\in \{521, \ldots, 565\}$

\subsubsection{The phi-level region set instance weighting function}
In summary, according to Eqs. \eqref{equ:p_zero_cum}, and \eqref{equ:update}, the incremental weighting function of the core instance $d_i$  of a region $\mathcal{B}_i^{\phi}(d_i, \varepsilon_i^\phi)$ is defined as
\begin{equation}
\label{equ:weight_update}
  w_{\phi}^\tau(d_i) =
    \begin{cases}
      1, & if ~ \|X_\tau - X_i \| \leq \varepsilon_i^\phi \\
      (1-\phi)\cdot w_{\phi}^{\tau-1}(d_i), & \text{otherwise}
    \end{cases}
\end{equation}
where $w_{\phi}^0 (d_i)=1$.

In a concept drift adaptation scenario, the region $\mathcal{B}_i^{\phi}(d_i, \varepsilon_i^\phi)$ will be removed from $\mathbb{B}_{\phi}$ if $w_{\phi}^{\tau}(d_i)<\alpha$. Also, the region with the lowest weight will be replaced by the latest data instance and its region if the region set size reaches a predefined threshold, called maximum buffer size. denoted as $w_{\max}$. A $\phi$-level ensembler is trained based on the core instance set with the weights from the $\phi$-level region set. That is, the training set is $\mathbb{B}_\phi(D)=\{d_i:\mathcal{B}_i^{\phi}(d_i, \varepsilon_i^\phi) \in \mathbb{B}_{\phi} \}$.

\subsubsection{A strategy to control the impact of false alarms}

As argued by Tsymbal et al. \cite{T:local_IF}, region drift should be defined as changes in concepts (data distributions) at the instance level, not at the dataset level. Therefore, to avoid detecting redundant drifts, and to mitigate the impact of false alarms, only the core instance of a region should be updated/removed. Since each region has a unique core instance, the weights of the overlapped instances will not change, and the weight of the core data instances can be incrementally adapted based on the risk of drift in their region.

\subsection{A maximum region drift disagreement ensemble }
\label{ss:max-diversity}

\subsubsection{Region drift disagreement}
Given two region set parameters $\phi_1$, $\phi_2$, and a training set $D_0$, we can build two region sets, $\mathbb{B}_{\phi_1},\mathbb{B}_{\phi_2}$. Without concept drift, we assume the data instances have arrived i.i.d. in an online manner. The RDD index of two region sets is defined as the Jaccard dissimilarity of the set of core instances for those regions, denoted as
\begin{equation}\label{equ:rdd}
    \mathrm{RDD}(\mathbb{B}_{\phi_1},\mathbb{B}_{\phi_2}) = 1-\frac{\mathbb{B}_{\phi_1}(D)\cap \mathbb{B}_{\phi_2}(D)}{\mathbb{B}_{\phi_1}(D)\cup \mathbb{B}_{\phi_2}(D)}.
\end{equation}
The RDD ensemble diversity is then defined as the average RDD of all pairs of ensemblers in the set of regions $\mathfrak{B}_{\Phi}=\{\mathbb{B}_{\phi_1},\ldots,\mathbb{B}_{\phi_i}\}$, formulated as
\begin{equation}\label{equ:diversity}
    \mathcal{D}(\mathfrak{B}_{\Phi})=\frac{2}{|\mathfrak{B}_{\Phi}|\cdot(|\mathfrak{B}_{\Phi}|-1)} \sum_{j=1}^{|\mathfrak{B}_{\Phi}|-1} \sum_{k=j+1}^{|\mathfrak{B}_{\Phi}|} \mathrm{RDD}(\mathbb{B}_{\phi_j},\mathbb{B}_{\phi_k}).
\end{equation}
Since $\phi$ determines the radius $\varepsilon$ for all regions in a region scheme, we can approach the power set of the feature space if there are sufficient data and $\phi$ covers all possible region sample proportion values. However, given there is always some level of constraint on computation costs, $\Phi$ cannot be infinite in real-world applications. Therefore, given that $\Phi$ is a set of grid values between 0 and 1, we want to select a subset of $\Phi$ with a limit number of $\varphi$ to reach the maximum RDD diversity (max-RDD), namely
\begin{equation}\label{equ:max-rdd}
    \mathrm{maxRDD}=\mathop{\max}_{\mathfrak{B}_{\Phi^{\prime}}\subseteq \mathfrak{B}_{\Phi}, |\mathfrak{B}_{\Phi^{\prime}}|=\varphi} \mathcal{D}(\mathfrak{B}_{\Phi^{\prime}}).
\end{equation}
where parameter $\varphi\leq|\Phi|$ governs how many ensemblers should be used for the final prediction. The goal of maximum ensemble diversity selection is to select a subset of region sets $\mathfrak{B}_{\Phi^{\prime}}\subseteq\mathfrak{B}_{\Phi}$ so that the diversity $\mathcal{D}$ reaches the maximum value. Many ensemble algorithms use 10 ensemblers as the default setting \cite{Bifet:ADWIN_ARF,Brzezinski:OAUE}, so we set $\varphi=10$ as default as well. The grid search range of $\Phi$ is set to $\{0.025,0.05,\ldots,0.5\}$ as a default. In this case, the max-RDD will select 10 out of 20 region sets from which to build the  base classifiers for making classifications

\subsubsection{The ensemble-voted classifier}
The ensemble-voted classifier is a meta-classifier that combines either similar or conceptually different machine learning classifiers for classification via majority voting. Two voting strategies can be used: "hard" and "soft". In hard voting, the final class label prediction is the most-frequently predicted class label by the classification models. In soft voting, the final class label is predicted by averaging the class-probabilities. For example, suppose we have three ensemble classifiers that have calculated their prediction probabilities on a binary classification task as $\{0.6, 0.4\}$, $\{0.7, 0.3\}$, $\{0.1, 0.9\}$, respectively. The combined prediction probability with hard voting would be $\{2, 1\}$; hence, the final prediction result would be the first class. With soft voting, the combined prediction probability would be $\{1.4, 1.6\}$, resulting in a final prediction of the second class. Soft voting is recommended if the classifiers are well-calibrated. Since the voting ensemblers in DiwE are selected via max-RDD, we want to leverage the advantage of soft voting to reflect their disagreement on the combined result. Thus, we chose soft majority voting as our final prediction strategy.

\begin{figure}[!ht]
    \centering
    \includegraphics[scale=0.8]{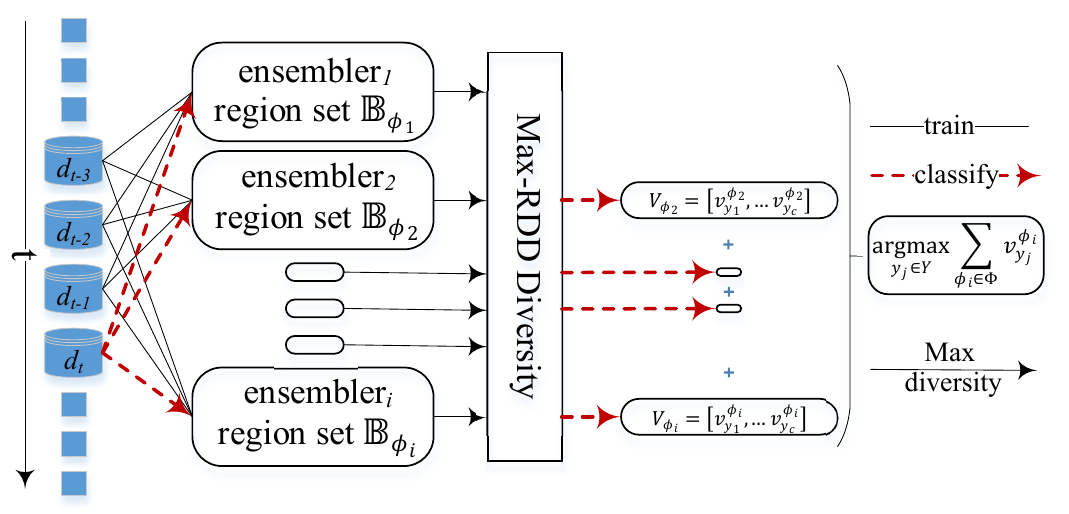}
    \caption{The structure of maximum RDD diverse instance weighting ensemble. Each $\phi_i$ is used to construct a region set $\mathbb{B}_{\phi_i}$ so that all regions in the set have an expected region sample proportion equal to $\phi_i$. A $\phi_i$-classifier, or selected $\phi_i$-ensemblers, are built based on the core instances of the region set. A total of $\varphi$ ensemblers are selected based on the max-RDD diversity. For each ensembler, the classification result is a labeled probability vector, denoted as $V_{\phi_i}=\{v_{y_1},\ldots,v_{y_c}\}$. $\Sigma_{j=1}^{c}v_{y_j}=1$, where $c$ is the number of classes. The final classification result is determined by a soft majority vote.}
    \label{fig:structure}
\end{figure}

Recall that the fundamental idea of a max-RDD ensemble-voted classifier is to select the most controversial drift detection region sets for ensemble learning. The ensemble determines whether two region sets are inconsistent according to an RDD index. If two region sets have no argument about the drift detection result, it is enough to preserve only one. However, if one detects a drift and the other does not, both need to be preserved for classification to boost the diversity.
The soft majority vote will balance the final classification result, which is
\begin{equation}\label{equ:majorityVoting}
    \mathrm{softMajorityVote}=\mathop{\arg\max}_{y_j\in Y} \Sigma_{\phi_i\in\Phi^{\prime}} v_{y_j}^{\phi_i}.
\end{equation}
where $y_j$ is the class label, and $v_{y_j}^{\phi_i}$ is the classification probability given by the base learner trained on the core instance set $\mathbb{B}_{\phi_i}(D)$. The structure of max-RDD DiwE is illustrated in Fig. \ref{fig:structure}.

It is worth to mention that different base classifier will calculate the $v$ differently. In this paragraph, we present formulation of $v$ in terms of IBk classifier. The $v$ of each IBk base learner is calculated as follows according to\cite{aha1991instance}:
\begin{itemize}
\item The algorithm parses the entire time window, computing the distance between $d_\tau$ and each training observation. The $k=5$ points in the training data that are closest to $d_\tau$ are denoted as the set $D_\tau ^ {knn}$.

\item Then the conditional probability for each class is estimated, i.e., the fraction of the number of points in $D_\tau ^{knn}$ with that given class label. Binary classification problems are given a class label ${c_0,c_1}$, and
\begin{equation*}
\begin{split}
    & P(y=c_0 |X=X_\tau) = \frac{\sum_{d_i\in D_\tau ^{knn}} I_{y_i=c_0}}{|D_\tau ^{knn}|} \\
    & P(y=c_1 |X=X_\tau) = \frac{\sum_{d_i\in D_\tau ^{knn}} I_{y_i=c_1}}{|D_\tau ^{knn}|}
\end{split}
\end{equation*}
\end{itemize}
where $d_i$ is a feature vector paired with a class label, $d_i=(X_i,y_i )$. The $I_{y_i=c_j}$ is an indicator function: if $y_i=c_0$ then $I_{y_i=c_0}=1$, otherwise $I_{y_i=c_0}=0$. Therefore, the probability vector $v={P(y=c_0 |X=X_\tau),P(y=c_1 |X=X_\tau )}$. The inverse-distance-weighted IBk consider the distance between samples as the weight to adjust the probability vector\mbox{\cite{Viktor:SAMkNN}}. The prediction probabilities $v$ are calculated differently for different base classifiers. However, considering drift detection occurs at the instance level, we recommend using the IBk classifier as the default.


\subsection{The implementation of the DiwE algorithm}
\label{ss:DiwE-implementation}

\subsubsection{The initialization of a region set}
To initialize a region set, we need a training dataset $D_0$ and a region set parameter $\phi_i$. The size of the training set is denoted as $m_0$. If the training set is not large enough, that is $m_0 < \max\{ 10/\phi_i, 10/(1-\phi_i)\}$, as discussed in Section \ref{subsub_m0}, a very large value is assigned to the radius. Therefore, the next data instance to arrive will definitely be located in this region. The region updating process is then triggered during which the radius is recalculated. This process continues until there are enough data instances.

\begin{algorithm}[ht]
    \caption{$\phi$-level Region Set Initialization}
    \label{alg:ini-region}
    \small
    \SetKwInOut{Input}{input}
    \SetKwInOut{Output}{output}

    \Input{1. the region set parameter, $\phi_i$
        \newline 2. training dataset, $D_0$
        }
    \Output{a $\phi$-level region set, $\mathbb{B}_{\phi_i}$}
    \BlankLine
    initial training dataset size, $m_0=|D_0|$\;
    initial the $\phi$-level region set, $\mathbb{B}_{\phi_i}=\{\}$\;

    \For {$d_t$ in $D_0$} {
        
        \eIf{$m_0 < \max\{ 10/\phi_i , 10/(1-\phi_i)\}$}{
            create region $\mathcal{B}_t^{\phi_i}=(d_t,$ Double.max$)$
            \tcp*[r]{if the training dataset is not large enough, we do not estimate the region radius.}
        }{
            estimates the $k$ value, $k=\lceil\phi_i\times m_0\rceil$\;
            find the $k$-nearest neighbour of $d_t$ in $D_0$, $d_k$\;
            estimates the region radius $\varepsilon_t^{\phi_i} = \|X_t - X_k \|$\;
            create region $\mathcal{B}_t^{\phi_i}=(d_t, \varepsilon_t^{\phi_i})$\;
        }
        $\mathbb{B}_{\phi_i}=\mathbb{B}_{\phi_i}\cup \{ \mathcal{B}_t^{\phi_i}\}$\;
    }
    \Return $\mathbb{B}_{\phi_i}$\;
\end{algorithm}
In the worst case of \hl{Algorithm \ref{alg:ini-region}}, the runtime complexity of each $k$-nearest neighbor search is $\mathcal{O}(m_0 n)$, where $n$ denotes the dimensionality. Given a fixed-size training set of $m_0$, the worst-case runtime complexity is $\mathcal{O}(m_0^2 n)$.

\subsubsection{Max-RDD diversity ensembler selection}
The intuition behind Max-RDD diversity is to quantitatively measure the disagreement between ensemblers about whether region drift exists, and only select the ensemblers that do not reach a consensus for the final prediction. As such, region drift is empirically estimated at the instance level.

In Algorithm \ref{alg:max-rdd}, the inputs are the set of region sets $\mathfrak{B}_{\Phi}$, and the maximum number of ensemblers for ensemble learning is $\varphi$. The RDD index is calculated in Line 9. The intersection and union runtime complexity are both $\mathcal{O}(2\delta)$, where $\delta$ denotes the cardinality of the region set.
The RDD index between two given region sets has a runtime complexity of $\mathcal{O}(2\delta + 2\delta)$
Hence, the complexity of calculating the RDD index for all pairs of region sets is $\mathcal{O}(|\Phi|^2(2\delta + 2\delta))$, $\mathcal{O}(4 |\Phi|^2 \delta)$.
To iterate over all the possible combinations of the region set parameters with a given voting ensemble size $\varphi$, we have $|\mathrm{Comb}_{\Phi}^{\varphi}|$ options, where $\mathrm{Comb}_{\Phi}^{\varphi}$ stands for all the possible combinations of choosing $\varphi$ out of $\mathfrak{B}_{\Phi}$ region sets. The cardinality of $\mathrm{Comb}_{\Phi}^{\varphi}$ is calculated by the combination function $C_n^r$. In this case, the cardinality is $C_{|\Phi|}^{\varphi}$, so the runtime complexity is $\mathcal{O}(C_{|\Phi|}^{\varphi})$.
The overall complexity of Max-RDD is $\mathcal{O}(4|\Phi|^2 \delta + C_{|\Phi|}^{\varphi})$, where $\delta\in Z^+$ is the stored region set size, and $10/\phi_i \leq\delta\leq w_{\max}$ according to the buffer size constraints. The complexity of Max-RDD is independent of the size of the dataset.

\begin{algorithm}[ht]
    \caption{Max-RDD Diversity Ensembler Selection}
    \label{alg:max-rdd}
    \small
    \SetKwInOut{Input}{input}
    \SetKwInOut{Output}{output}

    \Input{1. a set of region sets $\mathfrak{B}_{\Phi}$
        \newline 2. voting ensemble size (default $\varphi=10$)
        }
    \Output{Max-RDD diversity selected region sets $\mathfrak{B}_{\Phi^{\prime}}^{\max}$}
    \BlankLine
    initial $\mathrm{RDD_{aver}}=0$\;
    initial $\mathrm{RDD_{max}}=-1$\;
    find all combinations $\mathrm{Comb}_{\Phi}^{\varphi} = \mathrm{Comb}(\mathfrak{B}_{\Phi}, \varphi)$\;
    initial $\mathfrak{B}_{\Phi^{\prime}}^{\max}=\mathrm{Comb}_{\Phi}^{\varphi}[1]$ \tcp*[r]{the first item}

    \For {$\mathfrak{B}_{\Phi^{\prime}_i}$ in $\mathrm{Comb}_{\Phi}^{\varphi}$} {

        $\mathrm{RDD_{aver}}=0$\;
        \For {$j$ in range$\big(1, |\mathfrak{B}_{\Phi^{\prime}_i}|-1\big)$}{

            \For{$k$ in range$\big(j+1, |\mathfrak{B}_{\Phi^{\prime}_i}|\big)$}{
                $\mathrm{RDD_{aver}}=\mathrm{RDD_{aver}}+\mathrm{RDD}(\mathbb{B}_{\phi_j},\mathbb{B}_{\phi_k})$, where $\mathbb{B}_{\phi_j}, \mathbb{B}_{\phi_k}\in \mathfrak{B}_{\Phi^{\prime}_i}$ \tcp*[r]{Eq. \eqref{equ:rdd}}
            }
        }

        $\mathrm{RDD_{aver}}=\frac{2\cdot \mathrm{RDD_{aver}}}{|\mathfrak{B}_{\Phi^{\prime}_i}|\cdot(|\mathfrak{B}_{\Phi^{\prime}_i}|-1)}$ \tcp*[r]{Eq. \eqref{equ:diversity}}
        \If{$\mathrm{RDD_{aver}} > \mathrm{RDD_{max}}$}{
            $\mathrm{RDD_{max}}=\mathrm{RDD_{aver}}$\;
            $\mathfrak{B}_{\Phi^{\prime}}^{\max} = \mathfrak{B}_{\Phi^{\prime}_i}$\;
        }

    }
    \Return $\mathfrak{B}_{\Phi^{\prime}}^{\max}$
\end{algorithm}

\subsubsection{Diverse instance weighting ensemble}


\begin{algorithm}[ht]
    \caption{Diverse instance weighting ensemble}
    \label{alg:DiwE}
    \small
    \SetKwInOut{Input}{input}
    \SetKwInOut{Output}{output}

    \Input{1. training dataset $D_0$
        \newline 2. data stream $d_1,\ldots,d_m$
        \newline 3. region set parameter set (default $\Phi=\{0.025,0.05,\ldots,0.5\}$)
        \newline 4. voting ensemble size (default $\varphi=10$)
        \newline 5. base learner, $L$ (default IBk Classifier, $k=5$, weighting: inverse distance)
        \newline 6. max buffer size, $w_{\max}$ (default $1000$)
        \newline 7. drift significance level, $\alpha$ (default 0.01)
        }
    \Output{prediction results, $d_1^{\hat{y}},\dots,d_n^{\hat{y}}$}
    \BlankLine
	\For{$\phi_i$ in $\Phi$}{
	    initial region set $\mathfrak{B}_{\Phi}=\mathfrak{B}_{\Phi}\cup\{\mathbb{B}_{\phi_i}\}$ based on $D_0$\;
	}
    \While{stream not end, denote current time as $\tau$}{

        $\mathfrak{B}_{\Phi^{\prime}}=\mathrm{MaxRDD}(\mathfrak{B}_{\Phi},\Phi , \varphi)$\;

        \For{$\mathbb{B}_{\phi_i}$ in $\mathfrak{B}_{\Phi^{\prime}}$}{
	       train ensembler $L_{\phi_i}=\mathrm{Train}\big(\mathbb{B}_{\phi_i}(D)\big)$, the ensemble is $\mathcal{L}_{\Phi^{\prime}}=\mathcal{L}_{\Phi^{\prime}}\cup\{L_{\phi_i}\}$\;
	    }
        $d_{\tau}^{\hat{y}}=$softMajorityVote$(\mathcal{L}_{\Phi^{\prime}},d_{\tau})$ \tcp*[r]{Eq. \eqref{equ:majorityVoting}}

        \For{$\mathbb{B}_{\phi_i}$ in $\mathfrak{B}_{\Phi}$}{

            \For{$\mathcal{B}_j^{\phi_i}$ in $\mathbb{B}_{\phi_i}$}{

                update weight $w_{\phi_i}^{\tau}(d_j)$ \tcp*[r]{Eq. \eqref{equ:weight_update}}

                    \If{$w_{\phi_i}^{\tau}(d_j)<\alpha$}{
                        remove $\mathcal{B}_j^{\phi_i}$\;
                   }
            }
            add new region $\mathcal{B}_{\tau}^{\phi_i}$, $\mathbb{B}_{\phi_i}=\mathbb{B}_{\phi_i}\cup\{\mathcal{B}_{\tau}^{\phi_i}\}$\;

            \If{$|\mathbb{B}_{\phi_i}|>w_{\max}$}{
                Remove $\mathcal{B}_z^{\phi_i}=\underset{\mathcal{B}_j^{\phi_i}\in \mathbb{B}_{\phi_i}}\min w_{\phi_i}^{\tau}(d_z)$ where $d_z \in \mathbb{B}_{\phi_i}(D)$\;
            }
        }
    }
    \Return $d_1^{\hat{y}},\dots,d_m^{\hat{y}}$
\end{algorithm}

The idea behind DiwE is to use a buffer to store the regions that are most relevant to the current concept and to update the learning models regularly according to the core instances in the stored regions, i.e., $\mathbb{B}_{\phi}(D)$. We set $\Phi=\{0.025,0.05,\ldots,0.5\}$ so that $|\Phi|=20$ by default, which means there are 20 different $\phi$ values. Then we selected 10 as representatives, to vote on the final prediction result using the Max-RDD ensembler selection algorithm. The number of representatives selected is determined by a voting ensemble size parameter, denoted as $\varphi=10$. In general, we manipulated 20 ensemblers at all times and dynamically chose 10 of them for voting.
The configuration of $\Phi$ should be selected according to the available computational resources. 
The larger the $\Phi$ size, the more different region sets DiwE can investigate. The downside is that a larger $\Phi$ will increase the algorithm complexity in a combinatorial manner. Therefore, we recommend determining the size of $\Phi$ according to the computational resources, then fill in the values by a grid searching from $0$ to $0.5$.
Considering DiwE detect drift at the instance level, we recommend using the IBk classifier as the default.
The input parameter $w_{max}$ indicates the maximum number of regions that are allowed to be stored in a region set. In common practice, the default setting of $w_{max}$ is $1000$ \cite{Liu:PR,Viktor:SAMkNN}.

In Algorithm \ref{alg:DiwE}, the system is initialized on the training data in Lines 1-3. Line 4 starts processing the streaming data. Line 5 selects the ensemblers according to max-RDD diversity. The base learners are built in Lines 6-8, and Line 9 is the soft majority vote for the new data instance following Eq. \eqref{equ:majorityVoting}. Lines 11-16 track region drift and update instance weight according to Eq. \eqref{equ:weight_update}.
Line 17 is where the new regions are constructed as new data becomes available. Similar to the region initialization algorithm, we find the $k=\lceil\phi_i\times \delta\rceil$-th nearest neighbor of $d_{\tau}$, and build the region with $\mathbb{B}_{\tau}^{\phi_i}=(d_{\tau}, \|X_\tau - X_k\|)$.
Lines 18-20 ensure the buffer size does not exceed the maximum limit by removing the most likely drift regions.
Lines 22 and 23 detect the end of the stream and output the prediction/classification results.

\begin{table}[!ht]
\centering
\caption{DiwE Runtime Complexity Summary}
\label{tab:alg_complexity}
\begin{tabular}{@{}lll@{}}
\toprule
Line Number                 & Function                         & Complexity                                           \\ \midrule
Line 2 & Region set construction          & $\mathcal{O}(m_0^2 n )$                              \\
Line 5                      & Max-RDD                          & $\mathcal{O}(4 |\Phi|^2\delta+C_{|\Phi|}^{\varphi})$ \\
Line 6-8                    & IBk classifier ensemble          & $\mathcal{O}(\varphi)$                               \\
Line 9                      & softMajorityVote                 & $\mathcal{O}( \varphi \delta n)$                     \\
Line 11-16                  & Weight updating                  & $\mathcal{O}(\delta n)$                              \\
Line 17                     & Creating a new region            & $\mathcal{O}(1)$                                     \\
Line 18-20                  & Least important instance removal & $\mathcal{O}(\delta)$                                \\ \bottomrule
\end{tabular}
\end{table}

In terms of complexity, the worst case for the region set construction (Lines 1-3) is $\mathcal{O}(|\Phi|m_0^2 n )$.
The complexity of Max-RDD is in Line 5 is $\mathcal{O}(4 |\Phi|^2\delta+C_{|\Phi|}^{\varphi})$.
The training complexity of the IBk classifier is $\mathcal{O}(1)$ because IBk classifier does not require a training process. \hl{The complexity of the for-loop between Line 6 and 8 is $\mathcal{O}(\varphi)$.}
The softMajorityVote complexity of $\varphi$ IBk classifiers in Line 9 is $\mathcal{O}( \varphi \delta n)$.
The complexity of calculating the distance between $d_j$ and $d_{\tau}$, then compare the distance with $\varepsilon_{\tau}^{\phi_i}$ to update the weights is $\mathcal{O}(n)$. Updating the weights for a region set (Lines 11-16) is $\mathcal{O}(\delta n)$.
In Line 17, the complexity for creating a new region after computing the distance to all data instances in the region is $\mathcal{O}(1)$, given the calculations in Lines 11-16.
Removing the least important instance in Lines 18-20 is $\mathcal{O}(\delta)$ based on a minimum value search iteration of the buffer. So, the overall complexity for Lines 10-21 is $\mathcal{O}(|\Phi| ( \delta n + 1  + \delta ))$ Extending this to Lines 1-22, we have runtime complexity of $\mathcal{O}(|\Phi|m_0^2 n$$+$$m(4|\Phi|^2 \delta$$+$$C_{|\Phi|}^{\varphi}$$+$$\varphi$$+$$\varphi \delta n$$+$$|\Phi|(\delta n$$+$$1$$+$$\delta)))$. The details are summarised in Table \ref{tab:alg_complexity}

Simplified, this is
\begin{equation}\label{equ:complexitu}
    \mathcal{O}(\underbrace{|\Phi|m_0^2 n}_\text{$\mathrm{item_1}$} + \underbrace{(|\Phi|^2 \delta +C_{|\Phi|}^{\varphi})m}_\text{$\mathrm{item_2}$} + \underbrace{(\varphi+|\Phi|)\delta mn}_\text{$\mathrm{item_3}$})
\end{equation}
where $\mathrm{item_1}$ is the region sets initialization algorithm, $\mathrm{item_2}$ is the Max-RDD algorithm, $\mathrm{item_3}$ is the drift adaption process for the IBk classifiers.
Note that the runtime complexity of Algorithm \ref{alg:DiwE} is controlled by the input parameters, $D_0$, $\Phi$, $\varphi$, and $w_\mathrm{max}$. If all parameters are set as default values, then the overall complexity is $\mathcal{O}(\delta mn)$, where $\delta$ is the buffer size, and the worst case is $\delta=w_\mathrm{max}$, therefore, the complexity is $\mathcal{O}(w_\mathrm{max}  mn)$ which is similar to most stream learning algorithms.

%
%
%

\subsubsection{A scalability analysis and the data pre-processing requirements for DiwE}
Scalability is a system's capacity for handling a growing amount of work by adding resources to the system \cite{Bondi:scalability}, and is an important property in data stream learning algorithms. Resources fall into two broad categories: horizontal and vertical \cite{4228359}. From a data perspective, horizontal resources are the number of features ($n$), and vertical resources are the number of training/testing samples ($m$). A common way to increase the vertical scalability of an algorithm is sub-sampling - that is, using bagging or boosting algorithms to select a relatively small subset of samples to build the learning model. In time series and data stream mining tasks, variable time windowing strategies are an alternative approach \cite{Liu:IJCNN2019, Liu:FuzzyWindow}. In terms of horizontal scalability, dimension reduction is the most popular way to reduce the number of random variables under consideration. Many feature selection and feature projection techniques have been developed to address scalability, such as principal component analysis and auto-encoders \cite{Boracchi:QTree}. These techniques are usually used as a pre-processing step followed by clustering using $k$-nearest neighbor on feature vectors in a reduced-dimension space. In machine learning, this process is sometimes called low-dimensional embedding \cite{shaw2009structure}.

In DiwE, the runtime complexity is closely related to the algorithm parameters, $D_0$, $\Phi$, $\varphi$, and $w_\mathrm{max}$. DiwE can fit large datasets by adjusting these parameters to suit the available system capacity - for example, by reducing the window size $w_\mathrm{max}$ to control memory costs. However, directly applying DiwE to data with high dimensionality could be dangerous. That may cause memory overflows and significantly increase runtime complexity. Hence, we recommend applying a dimension reduction process before applying DiwE to high-dimension data. Which dimension reduction technique is best to use depends on the dataset and learning task at hand.

Another important issue that may affect DiwE's performance the chosen distance metric. Euclidean distance may not be efficient when dealing with data with many nominal attributes. Therefore, data pre-processing is essential, and feature normalization with  one-hot encoding is recommended in most cases.


\section{Experiments and Evaluation}
\label{s:DiwE_evaluation}

This section contains the evaluations of the proposed DiwE algorithm on both synthetic and real-world datasets. In Section \ref{ss:DiwE_exp1}, we demonstrate how a single DiwE member incrementally adapts to concept drift. In Section \ref{ss:DiwE_exp2}, we outline the ten synthetic datasets with both drifting and non-drifting streams that were used to compare accuracy. Section \ref{ss:DiwE_exp3} includes seven real-world benchmark datasets, and an evaluation of the Max-RDD ensembler selection. Performance was measured as accuracy, and all the results were evaluated by a prequential, basic classification performance evaluator.

\subsection{An evaluation of DiwE members on drift instance removal}
\label{ss:DiwE_exp1}

We first assessed how well a single DiwE member maintains its region set. This experiment was designed to evaluate whether the buffer size changes with concept drift, and whether the reserved core data instances in each region convey information about the most recent concept. To illustrate how the adaptation works, we used sliding windows with the same buffer limitation as a contrast. We also applied the Kolmogorov-Smirnov two-sample test (KS test) as a baseline to illustrate the difference between DiwE and the conventional concept drift retrain procedure.

\textbf{Experiment 1.} (Evaluation of a single DiwE member on drift instance removal.)
The datasets were generated based on the three different distributions given in Table \ref{tab:exp1_char}. One data instance was independently generated for each time point according to the current distribution. To simulate sudden and incremental drift, the data distributions were incrementally changed for $t\in \{1251,\ldots ,1750\}$ and suddenly changed at $t=2500$.
To maintain the KS test data buffer, we applied the most commonly used drift adaptation strategy \cite{I:HDDM, Gama:DDM}, that is, building a new buffer at a specified warning level and replacing the old buffer at a specified drift level. The warning level was set as $\alpha_{\mathrm{warn}}=0.05$, and the drift level was set to $\alpha_{\mathrm{drift}}=0.01$.

\begin{figure*}[!ht]
    \centering
    \includegraphics[scale=0.62]{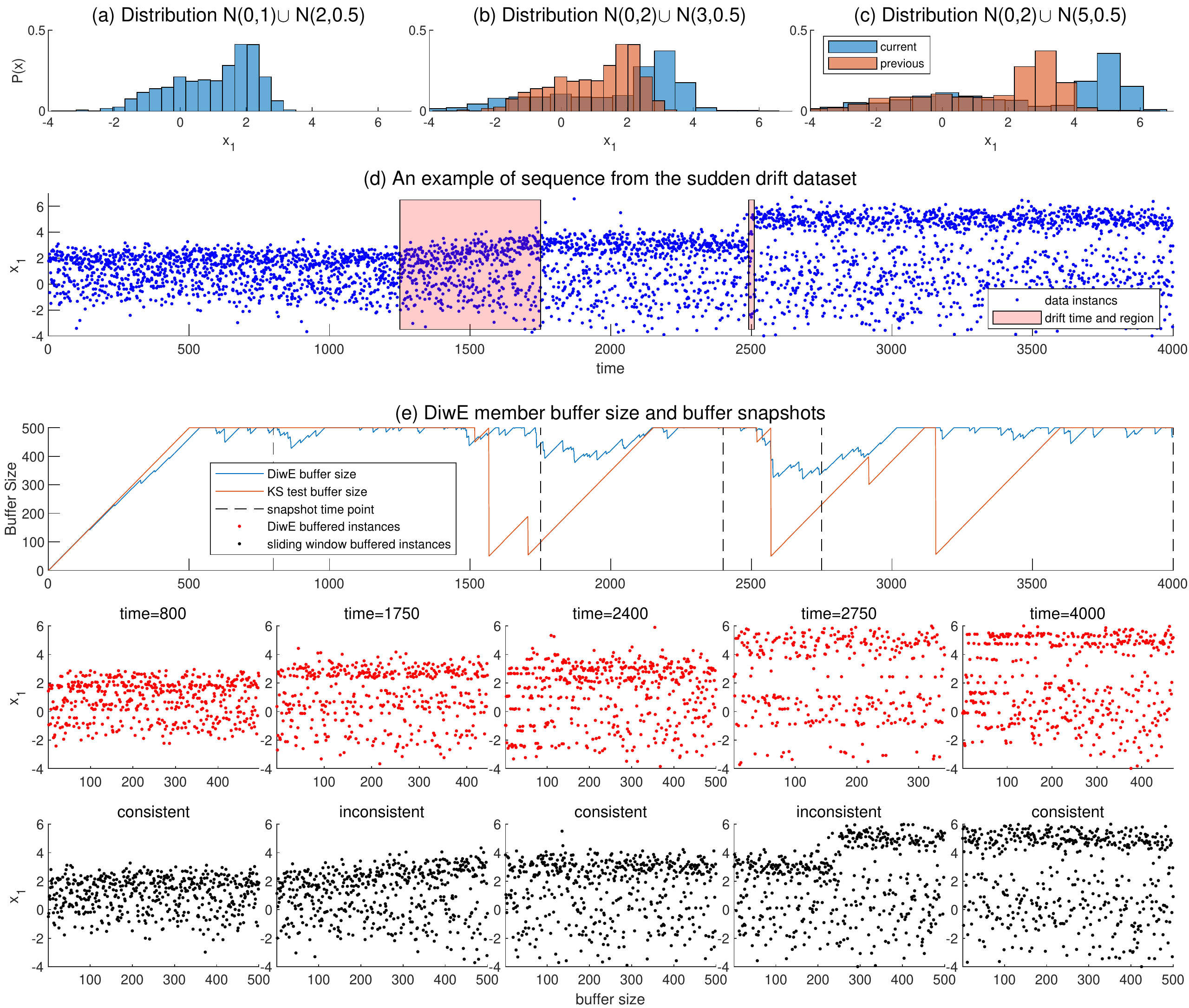}
    \caption{
    The experimental evaluation of DiwE on drift instance removal. Subfigures (a-c) show the empirical distribution of the data for the three different concepts. Subfigure (d) shows the data instances arriving at different times with incremental drift in $t\in\{1251,\ldots,1750\}$ and sudden region drifts at $t=2500$. Subfigure (e) demonstrates how the buffer size changes within evolving data streams. The 10 snapshots show the data instances stored in the buffer at different time points after both sudden and incremental drifts. At $t=800,2400,4000$, the system had already recovered from the drift; hence, the data instances are almost the same as the sliding windows. At $t=1750,2750$, the system was recovering, so the buffer size was reduced and the data instances stored in the buffer are different from the sliding windows. It also can be seen that the incremental drift changed more slowly than the sudden drift, so fewer instances were removed from the buffer than with sudden drift. Another problem raised in this experiment is that the warning level ($\alpha_{warm}=0.05$) was hardly triggered. A change detection test may skip the warning level and reach the drift level directly.}
    \label{fig:onlineRDA_exp1_result}
\end{figure*}

\begin{table}[!ht]
\centering
\caption{One-dimensional sudden-incremental drift data generator, where $\sigma_{\mathrm{inc}}$$=$$1+0.002\times (t-1250)$, $\mu_{\mathrm{inc}}$$=$$2+0.002\times(t-1250)$}
\label{tab:exp1_char}
\scriptsize
\begin{tabular}{@{}lll@{}}
\toprule
Drift type             &  Time slot          &  $x_1$ distribution                                                        \\ \midrule
no drift               &  $t\in\{1,\ldots,1250\}$     &  $x_1\sim N(0,1)\cup N(2,0.5)$                                             \\
incremental drift      &  $t\in\{1251,\ldots,1750\}$  &  $x_1\sim N(0,\sigma_{\mathrm{inc}}) \cup N(\mu_{\mathrm{inc}}, 0.5)$     \\
no drift               &  $t\in\{1751,\ldots,2500\}$  &  $x_1\sim N(0,2)\cup N(3,0.5)$                                             \\
sudden region drift  &  $t\in\{2501,\ldots,4000\}$  &  $x_1\sim N(0,2)\cup N(5,0.5)$                                             \\ \bottomrule
\end{tabular}
\end{table}

\textbf{Findings and Discussion}.
The experimental results are shown in Fig. \ref{fig:onlineRDA_exp1_result}, in a similar format to \cite{Alippi:HCDTs}. In general, both KS test and the DiwE member were able to take corrective actions no matter what type of drift occurred. However, we can see from the buffer size that the conventional replace and retrain method discarded all historical data after confirming a sudden drift, even though some of that data might still have been useful. The DiwE, in contrast, was able to trim irrelevant information from the buffer while retaining historical data that conformed to the current distribution. In addition, KS test triggered more than one true positive alarm during the incremental drift, which is correct from a drift detection perspective. However, the available training data in the buffer was overly reduced, which may not be necessary for drift adaptation. Compared to the sliding window strategy, DiwE is more sensitive to drift and can preserve the data instances that convey information about the most recent concept, as shown in the buffer snapshot at different time points. Another interesting result shown in this experiment is that KS test did not trigger a warning level but rather triggered a drift level directly on the incremental drift. This phenomenon inspired us to reconsider incremental drift as a series of sudden drifts. Notably, the warning level threshold of $\alpha=0.05$ may not always be the best option.

\subsection{An evaluation of DiwE on synthetic concept drift datasets}
\label{ss:DiwE_exp2}

In Experiment 2, we evaluated DiwE on ten synthetic datasets and compared it with eight state-of-the-art concept drift detection-adaptation algorithms.

\textbf{Experiment 2.} (Evaluation of DiwE on synthetic datasets) Synthetic datasets are good for generating and testing performance with specific and/or varied drift behaviors  \cite{Zliobaite:ActiveLearning, Gama:Evaluation
}. In this experiment, we applied seven data stream generators based on MOA \cite{B:MOA} with common parameterization \cite{Bifet:ADWIN_ARF, Yu:HLFR, Alippi:LSDD-CDT }. Table \ref{tab:exp2_char} shows the main characteristics of the datasets. The selected algorithms were ADWIN-ARF \cite{Bifet:ADWIN_ARF}, $\mathrm{LevBag_{kNN}}$ \cite{Bifet:ADWIN_LeverageBagging}, OnlineAUE \cite{Brzezinski:OAUE}, Learn++NSE \cite{Polikar:NSE}, SAMkNN \cite{Viktor:SAMkNN}, IBLStream \cite{Shaker:IBLStream}, and NN-$\mathrm{DVI_{kNN}}$ \cite{Liu:PR}, all of which are online mode classifiers. We ran the experiment using the MOA software framework to allow for easy reproducibility. Since different base classifiers may affect the results \cite{Liu:PR}, the base classifiers for $\mathrm{LevBag_{kNN}}$, Learn++NSE, SAMkNN, IBLStream, and NN-$\mathrm{DVI_{kNN}}$ were set as IBk, with a window size equal to 1000 and $k=5$. Neighbors were weighted by the inverse of their distance. ADWIN-ARF and OnlineAUE were only available with Hoeffding decision tree as the base classifiers. These two algorithms were selected because they are two benchmark ensemble algorithms for drift adaptation.

Similar to \cite{Bifet:ADWIN_ARF}, ten synthetic datasets were generated for evaluation: SEA sudden, gradual drift, Hyperplane incremental drift, LED sudden, gradual drift, AGR sudden, gradual drift, RTG no-drift, RBF global, and region drift. The characteristics of these datasets are summarized in Table \ref{tab:exp2_char}.

\begin{table}[!ht]
\centering
\caption{Characteristics of the ten synthetic datasets. (S: Sudden Drift, G: Gradual Drift, R: Region Drift)}
\label{tab:exp2_char}
\begin{tabular}{@{}lllll@{}}
\toprule
Dataset     & Drift Type    & \#Instances   & \#Attributes  & \# Class \\
\midrule
SEAs         & sudden        & 10k           & 3             & 2 \\
SEAg         & gradual        & 10k           & 3             & 2 \\
Hyp         & incremental   & 10k           & 10            & 2 \\
LEDs        & sudden                  & 10k           & 24            & 10 \\
LEDg        & gradual                 & 10k           & 24            & 10 \\
AGRs        & sudden                  & 10k           & 9             & 2 \\
AGRg        & gradual                 & 10k           & 9             & 2 \\
RTG         & no drift                & 10k           & 10            & 2 \\
RBF         & incremental             & 10k           & 10            & 5 \\
RBFr        & incremental, regional   & 10k           & 10            & 5 \\
\bottomrule
\end{tabular}
\end{table}

The \textbf{SEA} generator \cite{S:SEA} produces data streams with three continuous attributes, $X=\{x_1,x_2,x_3\}$ and $x_1,x_2,x_3\in[0,10]$. An inequality determines the label of each data instance, $x_1+x_2\leq\theta$, where $\theta$ is a threshold to control the label boundary. The entire data stream was divided into four subsets with different data distributions (“Concepts”) of equal size, and $\theta$ was 8, 9, 7, and 9.5, respectively. This evaluation method has been widely used in sudden drift detection and adaptation \cite{Liu:FuzzyWindow, Yu:HLFR, Bifet:ADWIN_ARF, Xu:DELM}. There were 10,000 data instances at a noise ratio of 10\%. To simulate gradual drift, Concepts 1 and 2 were changed every 50 data instances from $t=2250$ to $t=2750$, that is, the data for $t\in\{2250,\ldots,2300\}$ was generated based on the new concept, while the data for $t\in\{2300,\ldots,2350\}$ was generated based on the old concept up to $t=2750$.

The rotating \textbf{Hyp}erplane generator \cite{Hulten:CVFDT} produces data streams with ten continuous attributes, $X=\{x_1,\ldots,x_{10}\}$ and $x_1,\ldots,x_{10}\in[0,1]$. The label boundary for classification was determined by $\Sigma_{i=1}^d w_i x_i\geq\theta$, where $d$ is the number of features related to drift, and $w_i$ are weights that randomly initialize in the range of $[0,1]$. Incrementally changing the threshold $\theta$ produces a rotating hyperplane label boundary, thereby generating incremental concept drifts. In this experiment we set $d=2$, that is, only the first two features had incremental drifts. Again, there were 10,000 data instances, and the noise ratio was set to $10\%$.

The \textbf{LED} \cite{Breiman:LED} generator creates instances with 24 Boolean features, but only seven features determine the class labels. The configurations to simulate four different concepts were as follows: the first three features were swapped  for $t\in\{2500,\ldots,5000\}$; the first five features were swapped for $t\in\{5000,\ldots,7500\}$; and  the first seven features were swapped for $t\in\{7500,\ldots,10000\}$. The gradual drift configuration was the same as the SEA gradual drift.

The \textbf{AGR}AWAL \cite{Agrawal:AGR} generator creates instances with six nominal and three continuous attributes. Ten functions are available to map instances into two classes. We used the first four functions in MOA to simulate four concepts of equal length. The same gradual drift configuration was applied to AGRg.

The Random Tree Generator (\textbf{RTG}) randomly builds a decision tree and randomly assigns a class label to each leaf node, after which the data is uniformly distributed to the leaf nodes. For this dataset, we applied the MOA default setting to create a non-drifting dataset.

The \textbf{RBF} generator creates data instances using a radial basis function. It creates centroids at random positions and associates them with a standard deviation value, a weight, and a class label. Incremental drifts are simulated by continuously moving the centroids. Both RBF and RBFr were parameterized with 50 centroids with a speed of change equal to 0.001. For the RBF incremental drift, 50/50 centroids are drifting, and for the RBF incremental region drift, 10/50 are drifting.

The evaluation results were calculated based on 50 runs of each dataset. The average accuracy and standard deviation of accuracy are given in Table \ref{tab:exp2_result}.

\textbf{Findings and Discussion}.

The results show that DiwE reached an average of rank 2.0 on the evaluated datasets, which sits at the top of all the algorithms. We conducted a Friedman test to determine whether the difference in results was significant and found a significant difference at $p<0.01$ From a further investigation of the difference between each pair with the Nemenyi post-hoc test, we found that only the difference between DiwE and Learn++. NSE was significant ($p<0.01$). All other pairs had a significance level above $0.01$.

Overall, the results show that DiwE was the most accurate on most datasets, with the exception of AGRa, and AGRg. This might be due to the distance metric used for constructing the regions, which in this case was Euclidean distance. Euclidean distance performed well on the normalized numerical datasets, but appeared to have difficulties with the datasets containing nominal attributes. We therefore recommend choosing the distance metric carefully according to the feature type in the dataset(s).

\begin{table*}[!ht]
\centering
\caption{
Prequential classification accuracy (\%). The value in brackets indicates the rank of the results in each row. The average rank (AvgRank) is the mean of the rank for each column.}
\label{tab:exp2_result}
\footnotesize
\begin{tabular}{@{}lllllllll@{}}
\toprule
           & DiwE     & $\mathrm{LevBag_{kNN}}$   & ADWIN-ARF        & $\mathrm{NN}$-$\mathrm{DVI_{kNN}}$       & IBLStream        & OnlineAUE                      & SAMkNN   & Learn++NSE  \\
\midrule
SEAs & 88.88$\pm$0.36(1) & 84.37$\pm$0.41(4) & 84.52$\pm$0.01(3) & 83.17$\pm$0.96(5) & 79.42$\pm$0.44(8) & 82.58$\pm$0.44(7) & 85.39$\pm$0.59(2) & 82.79$\pm$0.44(6) \\
SEAg & 88.76$\pm$0.33(1) & 84.20$\pm$0.60(3) & 84.15$\pm$0.22(4) & 82.70$\pm$0.70(5) & 79.37$\pm$0.46(8) & 82.34$\pm$0.43(7) & 85.04$\pm$0.40(2) & 82.64$\pm$0.43(6) \\
HYP  & 88.16$\pm$0.28(1) & 83.95$\pm$0.57(2) & 77.09$\pm$1.04(7) & 78.04$\pm$0.58(5) & 77.94$\pm$0.75(6) & 81.31$\pm$0.47(3) & 79.89$\pm$0.80(4) & 72.68$\pm$0.53(8) \\
LEDs & 72.07$\pm$0.42(1) & 70.18$\pm$0.88(2) & 63.31$\pm$1.32(4) & 56.20$\pm$0.53(6) & 56.98$\pm$1.03(5) & 66.80$\pm$0.58(3) & 52.12$\pm$0.52(7) & 49.54$\pm$0.61(8) \\
LEDg & 71.84$\pm$0.49(1) & 70.11$\pm$0.60(2) & 62.30$\pm$1.99(4) & 56.21$\pm$0.52(6) & 56.79$\pm$0.98(5) & 66.52$\pm$0.57(3) & 52.10$\pm$0.93(7) & 49.22$\pm$0.56(8) \\
AGRs & 85.41$\pm$0.76(6) & 88.41$\pm$0.66(4) & 90.34$\pm$1.99(2) & 89.62$\pm$0.23(3) & 90.53$\pm$0.27(1) & 87.02$\pm$0.85(5) & 49.12$\pm$0.91(8) & 52.85$\pm$1.20(7) \\
AGRg & 85.42$\pm$0.75(6) & 88.42$\pm$0.64(4) & 90.25$\pm$1.97(2) & 89.61$\pm$0.23(3) & 90.55$\pm$0.27(1) & 87.06$\pm$0.85(5) & 49.15$\pm$1.56(8) & 52.98$\pm$1.08(7) \\
RTG  & 88.69$\pm$1.51(1) & 73.94$\pm$4.75(4) & 74.68$\pm$0.59(3) & 72.34$\pm$4.50(5) & 65.79$\pm$5.62(6) & 76.41$\pm$4.40(2) & 59.59$\pm$5.66(8) & 64.10$\pm$3.20(7) \\
RBF  & 88.87$\pm$1.58(1) & 51.26$\pm$2.08(7) & 67.61$\pm$1.29(6) & 82.27$\pm$2.85(3) & 75.01$\pm$1.91(5) & 47.40$\pm$2.81(8) & 88.00$\pm$2.88(2) & 77.82$\pm$1.91(4) \\
RBFr & 92.12$\pm$1.12(1) & 54.66$\pm$4.86(8) & 78.13$\pm$2.35(6) & 89.99$\pm$1.47(3) & 82.78$\pm$1.62(4) & 56.69$\pm$4.49(7) & 90.99$\pm$1.74(2) & 79.69$\pm$1.52(5) \\
\midrule
AvgRank    & 2.0 (1)        & 4.0 (2)        & 4.1 (3)          & 4.4 (4)        & 4.9 (5)        & 5.0 (6.5)        & 5.0 (6.5)          & 6.6 (8)   \\
\bottomrule
\end{tabular}
\end{table*}

\subsection{An evaluation of DiwE on real-world applications}
\label{ss:DiwE_exp3}
In this set of experiments, we evaluated DiwE on real-world applications. Experiment 3 shows the buffer size of the ensembler using a region set parameter $\phi=0.1$. Experiment 4 shows the effectiveness of maximum diversity ensembler selection by comparing it with random ensembler selection. Experiment 5 evaluates the robustness of DiwE with different parameter settings.

\begin{figure*}[!ht]
    \centering
    \includegraphics[scale=0.5]{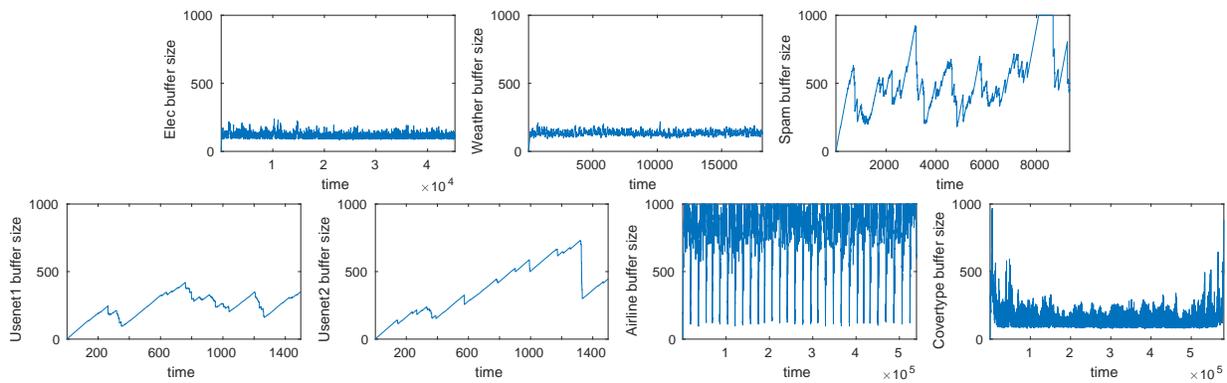}
    \caption{
    The buffer size of the region set $\mathbb{B}_{\phi=0.1}$ on real-world datasets. From the pattern of the buffer sizes, we conclude that the Electricity and Weather datasets frequently trigger drift alarms, which suggests that drift detection algorithms with high false alarm rates may also achieve good results. This issue is also discussed in [56]. The Spam, Usenet1, Usenet2, and Covtype datasets seem to have no specific drift patterns, while the Airline dataset seems to have regular pattern of drift.}
    \label{fig:exp3_result}
\end{figure*}

\textbf{Experiment 3.} (Evaluation of DiwE on seven real-world applications)
To evaluate the ability of DiwE to address real-world problems, we compared it with the same algorithms as introduced in Section \ref{ss:DiwE_exp2} but with real-world datasets. As discussed in \cite{Gama:Evaluation, Zliobaite:Evaluation, Bifet:Evaluation}, execution time and memory cost are important in streaming data learning, so this information has been provided alongside the results. The characteristics of the datasets used are summarized in \hl{Table \ref{tab:real_dataset_char}}. Tables \ref{tab:real_dataset_acc}, \ref{tab:real_dataset_time}, and \ref{tab:real_dataset_ram} show the performance of the tested algorithms, and Fig. \ref{fig:exp3_result} shows the changes in the size of the region set $\mathbb{B}_{\phi=0.1}$.

The \textbf{Elec}tricity dataset contains 45,312 instances, collected every 30 minutes from the Australian New South Wales Electricity Market between 7 May 1996 and 5 Dec 1998. In this market, prices are not fixed; rather, they are affected by supply and demand. This dataset contains eight features and two classes (up, down) and has been widely used to evaluate concept drift adaptation.

The Nebraska \textbf{Weather} prediction dataset was compiled by the US National Oceanic and Atmospheric Administration. It contains eight features and 18,159 instances with 31\% positive (rain) classes, and 69\% negative (no rain) classes. The dataset is summarized in \cite{Polikar:NSE} and is available at \cite{Polikar:Dataset}.

The \textbf{Spam} filtering dataset is a collection of 9324 email messages derived from the Spam Assassin collection and is available at http://spamassassin.apache.org/. The original dataset contains 39,916 features and 9324 emails. It is commonly considered to be a typical gradual drift dataset \cite{K:Spam500}. 
According to Katakis \cite{K:Spam500} 500 attributes can be retrieved using the Chi-square feature selection approach.

The \textbf{Usenet1} and \textbf{Usenet2} datasets are derived from Usenet posts in the 20 Newsgroup collection with simulated region drifts. The task is to classify messages as either interesting or junk as they arrive. The dataset is split into five periods, and the data in each period covers different user interest topics. All data instances were concentrated to simulate sudden/reoccurring drift. 

The \textbf{Airline} dataset consists of flight arrival and departure details for all commercial flights within the US from October 1987 to April 2008. The dataset was originally designed for regression problems as part of the Data Expo Competition, 2009. It was subsequently modified by the MOA team \cite{B:MOA} for prediction analysis. Each data instance has seven features and two classes with 539,388 records in total.

The forest cover type (\textbf{Covtype}) dataset designed to test predictions on the type of forest cover from a given observation as determined by the US Forest Service (USFS) Region 2 resource information system. Each instance is derived from data originally obtained from the US Geological Survey (USGS) and USFS data.

\begin{table}[!ht]
\centering
\caption{Real-world dataset characteristics}
\label{tab:real_dataset_char}
\begin{tabular}{@{}llll@{}}
\toprule
Dataset & \#Instances & \#Attributes & \#Class                          \\ \midrule
Elec    & 45312       & 8            & 2 (up, down)                     \\
Weather & 18159       & 8            & 2 (rain, no rain)                \\
Spam    & 9324        & 500          & 2 (spam, legitimate)             \\
Usenet1 & 1500        & 99           & 2 (interested, non-interested)   \\
Usenet2 & 1500        & 99           & 2 (interested, non-interested)   \\
Airline & 539383      & 7            & 2 (delay, not delay)             \\
Covtype & 581012      & 54           & 7 multiclass                     \\ \bottomrule
\end{tabular}
\end{table}

\begin{table*}[!ht]
\centering
\caption{Real-world dataset accuracy (\%).}
\label{tab:real_dataset_acc}
\small
\begin{tabular}{@{}lllllllll@{}}
\toprule
           & DiwE  & ADWIN-ARF  & $\mathrm{NN}$-$\mathrm{DVI_{kNN}}$  & $\mathrm{LevBag_{kNN}}$   & SAMkNN  & OnlineAUE  & IBLStream  & Learn++NSE  \\
\midrule
Elec    & 83.84 (5) & 88.17 (1) & 87.39 (3) & 84.35 (4) & 82.78 (6)  & 87.74 (2)  & 77.05 (7) & 69.67 (8) \\
Weather & 80.20 (1) & 78.74 (2) & 74.63 (7) & 74.97 (6) & 77.73 (3)  & 75.24 (5)  & 75.69 (4) & 73.24 (8) \\
Spam    & 96.69 (1) & 95.60 (3) & 93.47 (4) & 91.08 (6) & 95.79 (2)  & 84.29 (7)  & 92.78 (5) & 72.54 (8) \\
usenet1 & 68.53 (1) & 68.40 (2) & 66.80 (4) & 66.87 (3) & 65.67 (5)  & 63.47 (6)  & 56.00 (7) & 46.93 (8) \\
usenet2 & 73.20 (1) & 71.93 (4) & 72.47 (2) & 72.27 (3) & 71.00 (5)  & 68.87 (6)  & 67.67 (7) & 65.67 (8) \\
Airline & 78.55 (1) & 65.24 (4) & 64.55 (5) & 66.06 (3) & 60.35 (8)  & 67.51 (2)  & 63.74 (6) & 63.04 (7) \\
Covtype & 89.84 (6) & 92.11 (3) & 94.04 (1) & 84.74 (7) & 91.71 (4)  & 90.01 (5)  & 92.26 (2) & 68.43 (8) \\
\midrule
AveRank & 2.29 (1)  & 2.71 (2)  & 3.71 (3)  & 4.57 (4)  & 4.71 (5.5) & 4.71 (5.5) & 5.43 (7)  & 7.86 (8)  \\
\bottomrule
\end{tabular}
\end{table*}

\begin{table*}[!ht]
\centering
\caption{Execution time on real-world datasets (s CPU-Time). The time was computed based on a single ensembler}
\label{tab:real_dataset_time}
\small
\begin{tabular}{@{}lllllllll@{}}
\toprule
           & DiwE  & ADWIN-ARF  & $\mathrm{NN}$-$\mathrm{DVI_{kNN}}$  & $\mathrm{LevBag_{kNN}}$   & SAMkNN  & OnlineAUE  & IBLStream  & Learn++NSE  \\
\midrule
Elec    & 10.76    & 11.22      & 773.67     & 3.02        & 7.06     & 4.08      & 8.98      & 240.49     \\
Weather & 8.27     & 4.01       & 337.55     & 1.01        & 3.00     & 1.00      & 37.09     & 31.05      \\
Spam    & 15.73    & 6.01       & 1879.17    & 6.02        & 23.01    & 7.56      & 1131.02   & 1311.03    \\
usenet1 & 1.30     & 1.00       & 73.19      & 1.01        & 1.00     & 1.00      & 4.59      & 4.01       \\
usenet2 & 1.13     & 1.01       & 63.96      & 1.01        & 1.00     & 1.00      & 5.11      & 3.02       \\
Airline & 1867.97  & 355.19     & 8016.71    & 99.19       & 40.01    & 196.15    & 1.99      & 14657.84   \\
Covtype & 324.44   & 123.04     & 98.31      & 103.16      & 196.06   & 112.04    & 6432.80   & 102260.81  \\
\midrule
AveRank & 5.14 (5)    & 3.71 (4)   & 6.86 (7.5)  & 2.57 (2)    & 3.00 (3)   & 2.29 (1)   & 5.57 (6)    & 6.86 (7.5)       \\ \bottomrule
\end{tabular}
\end{table*}

\begin{table*}[!ht]
\centering
\caption{Memory cost on real-world datasets (GB RAM-Hours). The memory was computed based on a single ensembler}
\label{tab:real_dataset_ram}
\small
\begin{tabular}{@{}lllllllll@{}}
\toprule
& DiwE  & ADWIN-ARF  & $\mathrm{NN}$-$\mathrm{DVI_{kNN}}$  & $\mathrm{LevBag_{kNN}}$   & SAMkNN  & OnlineAUE  & IBLStream  & Learn++NSE      \\
\midrule
Elec    & 3.38E-03 & 1.19E-05   & 2.98E-08 & 1.01E-07    & 1.36E-05 & 5.79E-07  & 2.79E-06  & 7.64E-05   \\
Weather & 1.35E-03 & 4.54E-06   & 2.98E-08 & 4.91E-08    & 4.77E-06 & 1.68E-07  & 1.25E-05  & 3.91E-06   \\
Spam    & 6.95E-04 & 1.11E-05   & 2.89E-08 & 6.34E-06    & 1.83E-04 & 1.62E-05  & 6.86E-03  & 3.89E-03   \\
usenet1 & 1.12E-04 & 1.43E-07   & 2.70E-08 & 2.21E-08    & 8.14E-07 & 4.05E-09  & 1.85E-06  & 1.46E-06   \\
usenet2 & 1.12E-04 & 1.01E-07   & 2.70E-08 & 1.92E-08    & 9.59E-07 & 4.08E-09  & 2.08E-06  & 1.12E-06   \\
Airline & 4.02E-02 & 5.49E-03   & 6.30E-04 & 1.75E-03    & 8.41E-05 & 5.54E-03  & 2.14E-03  & 5.12E-01   \\
Covtype & 4.33E-02 & 1.61E-05   & 9.65E-02 & 9.92E-06    & 5.64E-04 & 1.56E-05  & 6.10E-03  & 4.16E+00   \\
\midrule
AveRank	& 7.29 (8)&	4.14 (4)	&2.57 (2)	&2.00 (1)	&4.57 (5)	&2.86 (3)&	6.00 (6)	&6.57 (7) \\ \bottomrule
\end{tabular}
\end{table*}

\textbf{Findings and Discussion}.
From the accuracy and execution efficiency results in Tables \ref{tab:real_dataset_acc}, \ref{tab:real_dataset_time}, and \ref{tab:real_dataset_ram}, we conclude that different drift adaptation algorithms are suited to different applications; there is no perfect algorithm that can achieve the best performance for all datasets. While the average ranking only demonstrates the effectiveness of DiwE on the tested datasets, the results do provide strong evidence that DiwE performs as well as the other methods in the tested situations. More concretely, what the results show is that considering diversity in region drift disagreement is a suitable alternative method for ensemble learning to address concept drift.

The memory cost of DiwE is higher than the other algorithms because a few region sets need to kept in memory. However, this issue could easily be overcome with distributed computing. From the results, we observe that the Covtype dataset has more attributes and data instances than the Airline dataset. But the execution times for DiwE, ADWIN-ARF, and NN-$\mathrm{DVI_{kNN}}$ on Covtype were much faster. From this, we surmise that the execution time of concept drift detection-adaptation algorithms might be related to the number of drifts in the dataset. Hence, differences in drift detection accuracy might result in different drift-adapt execution times. According to the complexity analysis of DiwE discussed in Section \ref{ss:DiwE-implementation}, DiwE has $\mathcal{O}(\delta mn)$ complexity, where $\delta$ denotes the buffer size. As shown in Fig. \ref{fig:exp3_result}, the average buffer size of the Airline dataset is much higher than the Covtype dataset, which accords with our conclusion. This phenomenon has inspired us to reconsider the balance between detection-adaptation performance and execution time.

\begin{table*}[ht]
\centering
\caption{Evaluation of the effectiveness of Max-RDD ensembler selection. We set $\varphi=10$, $\varphi=5$ and compared DiwE with random ensembler selection. The random DiwE ensembler selection results are summarised from 50 runs with mean and standard deviation. The values in the bracket are the difference between Max-RDD and random DiwE, which indicate Max-RDD is effective and the effectiveness increases as the $\varphi$ value decreases.}
\label{tab:max-rdd}
\begin{tabular}{llllllll}
\toprule
        & \multicolumn{3}{c}{$\varphi=10$}   &  & \multicolumn{3}{c}{$\varphi=5$}    \\ \cmidrule{2-4} \cmidrule{6-8}
        & \multicolumn{1}{c}{Max-RDD DiwE} &  & \multicolumn{1}{c}{Random DiwE}  &  & \multicolumn{1}{c}{Max-RDD  DiwE} &  & \multicolumn{1}{c}{Random DiwE}  \\
        \midrule
Elec    & 83.84   &  & 83.40$\pm$0.05 (0.44) &  & \textbf{84.96}   &  & 82.59$\pm$0.09 (2.37) \\
Weather & \textbf{80.20}   &  & 80.11$\pm$0.14 (0.09) &  & 79.93   &  & 78.82$\pm$0.21 (1.11) \\
Spam    & \textbf{96.69}   &  & 96.44$\pm$0.10 (0.25) &  & 96.58   &  & 78.82$\pm$0.21 (0.98) \\
usenet1 & \textbf{68.53}   &  & 67.92$\pm$0.72 (0.61) &  & 67.20   &  & 64.34$\pm$0.62 (2.86) \\
usenet2 & \textbf{73.20}   &  & 72.64$\pm$0.42 (0.56) &  & 69.93   &  & 69.32$\pm$0.89 (0.61) \\
Airline & 78.55   &  & 77.32$\pm$0.12 (1.23) &  & \textbf{79.28}   &  & 76.23$\pm$0.18 (3.05)  \\
Covtype & \textbf{89.84}   &  & 89.42$\pm$0.04 (0.42) &  & 89.36   &  & 88.42$\pm$0.03 (0.94)  \\ \bottomrule
\end{tabular}
\end{table*}

\textbf{Experiment 4.} (Evaluation of Max-RDD diversity ensembler selection.) To evaluate whether Max-RDD ensembler selection improves the overall classification results, we compared it with a random ensembler selection with the same $\Phi$ range. The aim of the Max-RDD ensembler selection is to select the most controversial region sets for ensemble learning so that the ensembles can reach a high drift sensitivity without losing robustness. Given this assumption, Max-RDD should be able to highlight the ensemblers with the highest diversity, no matter what type of drift, with a limited ensembler size $\varphi$. Random ensembler selection does not have this property, which means Max-RDD should outperform random ensembler selection. To verify our assumption, we chose $\varphi=10$, $\varphi=5$ and evaluated DiwE on the seven real-world datasets. For the random ensembler selection, we ran DiwE 50 times and calculated the mean and standard deviation. The results are shown in Table \ref{tab:max-rdd}.

\textbf{Findings and Discussion}. According to the Friedman test, there is a significant difference ($p<0.01$) in classification accuracy between the Max-RDD and random ensembler selection methods, but there is no significant difference for MaxRDD with different $\varphi$ values. The value in brackets indicates the extent to which Max-RDD improved classification accuracy compared to the random method. From the results, we see that a smaller $\varphi$ caused the random ensembler selection to become unstable, while Max-RDD maintained accuracy with no significant drops.
\begin{figure}[ht]
    \centering
    \includegraphics[scale=0.4]{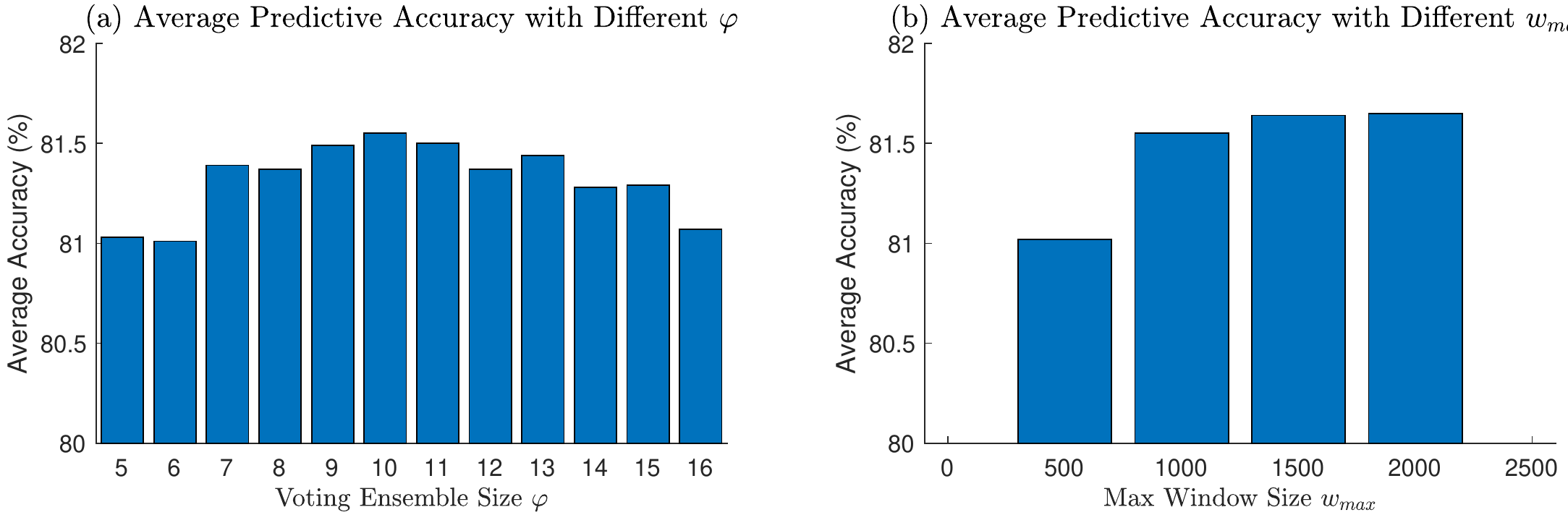}
    \caption{Evaluation results of the different voting ensemble size $\varphi$ (a), and different maximum window size $w_{\max}$ (b).}
    \label{fig:varphi}
\end{figure}

\textbf{Experiment 5.} 
(Evaluation of the selection of the voting ensemble size $\varphi$ and the maximum window size $w_{\max}$.) The voting ensemble size and the maximum window size are two critical parameters that may affect DiwE's performance. To evaluate how these parameters may influence prediction accuracy, we varied the settings of $\varphi$ and $w_{\max}$ with the real-world datasets. The summarized results are shown in Fig. \ref{fig:varphi}. To determine out how $\varphi$ impacts prediction, we set $\varphi$ in the range of 5 to 16 in steps of 1 and plotted the average accuracy (see Fig. \ref{fig:varphi}a). Then we set $\varphi = 10$ and changed the $w_{\max}$ from 500 to 2000 as shown in Fig. \ref{fig:varphi}(b).



\textbf{Findings and Discussion}. The Friedman test did not indicate a significant difference in the results with different values of $\varphi$ ($p<0.01$). However, we can see that, as $\varphi$ increased, average accuracy first increased and then decreased, with the highest average accuracy of 81.55\% at $\varphi = 10$. The lowest accuracy was 81.01\% at $\varphi = 6$. The results also show that $\varphi$ in the range 7 to 13 provided very similar accuracy. This phenomenon is reasonable. Accuracy is low when $\varphi$ is small because there are not enough ensemblers to join the voting process. But once there are sufficient ensemblers to provide a good level of diversity, the voting result improves (i.e., the RDD between each pair of ensemblers has a low variance). However, after a point, if $\varphi$ continues to increase, redundant ensemblers start to become involved in the voting process, which may jeopardize the soft voting strategy and accuracy begins to slide as a result. Consequently, one option for selecting the best $\varphi$ value could be trying all $\varphi$ values in the range of 1 to $|\Phi|$. This would significantly increase the runtime complexity and brute force methods are not very elegant. Instead, we recommend choosing $\varphi = \frac{|\Phi|}{2}$ as the default settings according to our empirical evaluation.


Regarding the maximum window size parameter $w_{\max}$, there was again no significant difference between the results with the selected window sizes. We can see that the accuracy increased from $w_{\max}=500$ to $w_{\max}=1000$ and remained stable over 1000. This might be due to the drift properties of the evaluated datasets. As DiwE will automatically remove drift instances, if the drift speed of the datasets is higher than the sample arrival speed. In this case, the system may not reach the maximum window size, and so the prediction accuracy will remain stable. The parameter $w_{\max}$ should be defined according to the available computational resources. If there is not enough memory, $w_{\max}$ should be reduced to fit the dataset.




\section{Conclusions and Future Work}
\label{s:VI}

In this paper, we introduced a novel region drift tracking method to serve as an online instance weighting function and presented an instance-based ensemble learning algorithm, called DiwE, to provide better concept drift adaptation results. The novelty of this research lies in its approach of measuring diversity according to the disagreements between ensemblers about the probability of region drift. The ensemblers with the most controversial differences can be selected dynamically and used to create diversity. The experimental results show that the overall performance of DiwE compares well with other state-of-the-art algorithms, indicating that region drift disagreement has potential as an alternative method of ensemble diversity measurement. We became aware, in this study, that incremental drift might cause drift detection algorithms to continually trigger drift alarms, which could have a negative impact on drift adaptation, and that the number of drifts in a dataset might increase an algorithm’s execution time. Therefore, a compromise may be required to balance execution time with drift detection-adaptation performance.

Based on our results, we plan to explore new methods of constructing regions in future research, such as introducing a fuzzy density clustering method or a more generalized region drift detection and adaptation algorithm. Further, since DiwE does not have a mechanism to store old concepts, it may also be desirable to model the drift instances DiwE removes as drifting clusters to allow for further investigation. Such a mechanism would be particularly useful for applications that deal with recurrent drift. Parallel and distributed concept drift detection-adaptation algorithms are another research direction. Lastly, we observed that the buffer size of DiwE may contain useful information about the characteristics of different types of drift, which could also be a worthy research topic.

\section*{Acknowledgments}
\hl{The work presented in this paper was supported by the
Australian Research Council (ARC) under Discovery Project DP190101733.} We also acknowledge the support of the NVIDIA Corporation with the donation of a GPU used for this research.

\ifCLASSOPTIONcaptionsoff
  \newpage
\fi



\bibliographystyle{jabbrv_ieeetr}
\bibliography{bib/reference}
%



%


\begin{IEEEbiography}[{\includegraphics[width=1in,height=1.25in, clip]{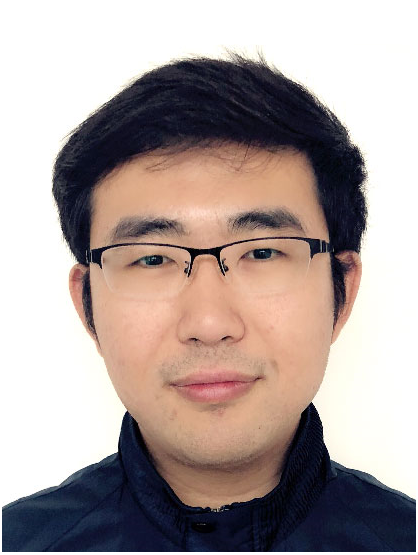}}]{Anjin Liu}
is a Postdoctoral Research Associate in the A/DRsch Centre for Artificial Intelligence, Faculty of Engineering and Information Technology, University of Technology Sydney, Australia. He received the BIT degree (Honour) at the University of Sydney in 2012. His research interests include concept drift detection, adaptive data stream learning, multi-stream learning, machine learning and big data analytics.
\end{IEEEbiography}

\begin{IEEEbiography}[{\includegraphics[width=1in,height=1.25in,clip,keepaspectratio]{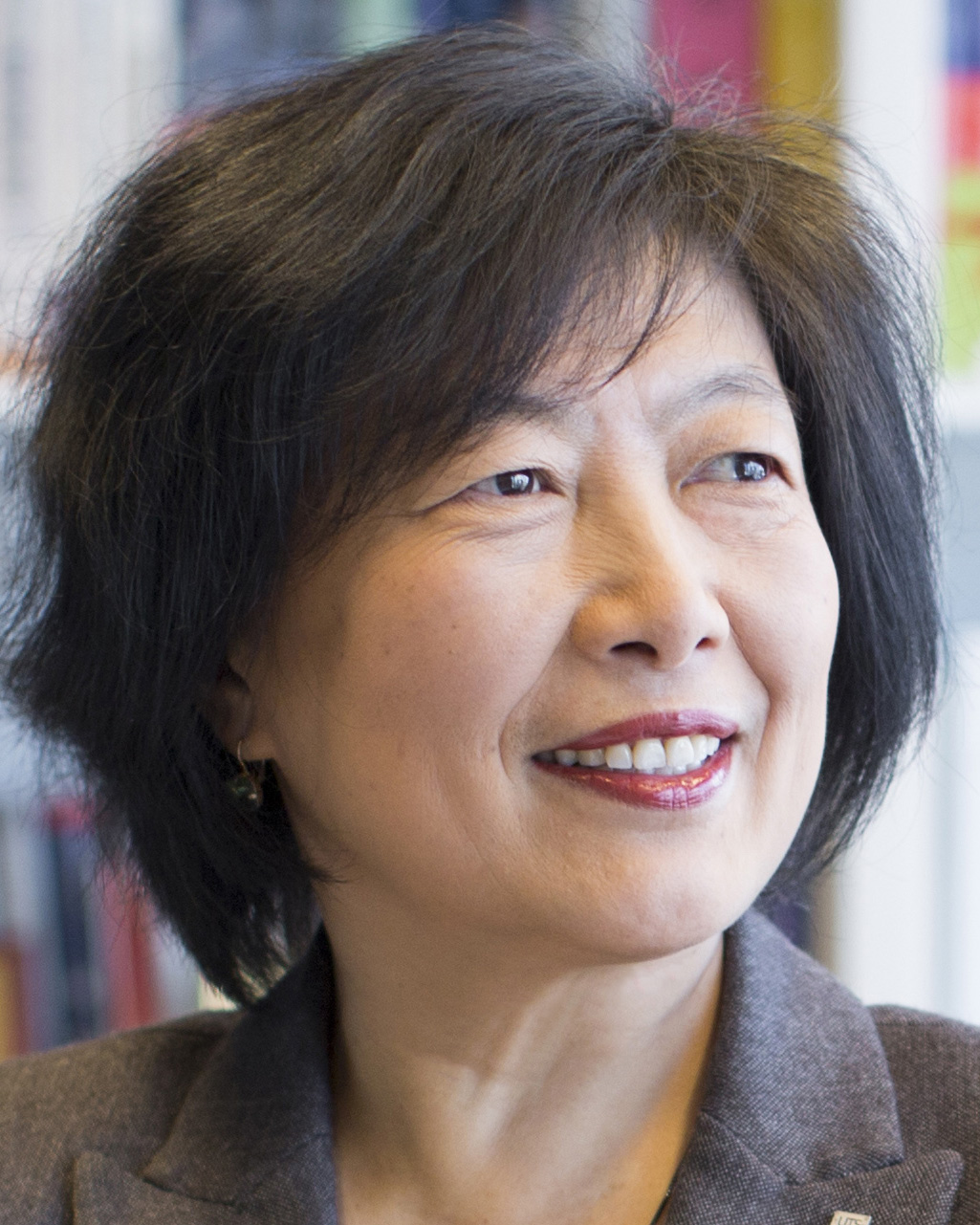}}]{Jie Lu}
(F'18) is a Distinguished Professor and the Director of the Centre for Artificial Intelligence at the University of Technology Sydney, Australia. She received the Ph.D. degree from Curtin University of Technology, Australia, in 2000. She is also a fellow of IFSA.

Her main research expertise is in fuzzy transfer learning, decision support systems, concept drift, and recommender systems. She has published six research books and 450 papers in IEEE Transactions on Fuzzy Systems and other refereed journals and conference proceedings. She has won over 20 Australian Research Council (ARC) discovery grants and other research grants. She serves as EditorIn-Chief for Knowledge-Based Systems (Elsevier) and Editor-In-Chief for International Journal on Computational Intelligence Systems (Atlantis), has delivered 25 keynote speeches at international conferences, and has chaired 10 international conferences. She has received awards of IEEE TFS outstanding paper award (2019) and Australian Innovative Engineer 2019 award.
\end{IEEEbiography}

\begin{IEEEbiography}[{\includegraphics[width=1in,height=1.25in, clip]{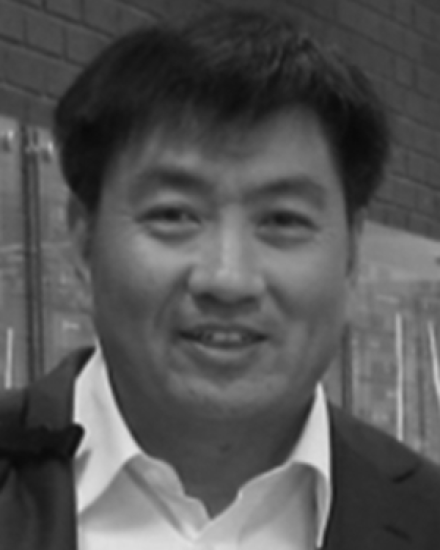}}]{Guangquan Zhang}
is an Associate Professor and Director of the Decision Systems and e-Service Intelligent (DeSI) Research Laboratory at the University of Technology Sydney, Australia. He received the Ph.D. degree in applied mathematics from Curtin University of Technology, Australia, in 2001.

His research interests include fuzzy machine learning, fuzzy optimization, and machine learning. He has authored five monographs, five textbooks, and  460 papers including 220 refereed international journal papers
Dr. Zhang has won seven Australian Research Council (ARC) Discovery Projects grants and many other research grants. He was awarded an ARC QEII fellowship in 2005.
He has served as a member of the editorial boards of several international journals, as a guest editor of eight special issues for IEEE transactions and other international journals, and co-chaired several international conferences and workshops in the area of fuzzy decision-making and knowledge engineering.
\end{IEEEbiography}




\end{document}